# Earth Embeddings in Environmental Remote Sensing: A Practitioner's Perspective

Yuchi Ma[1*], Jiaqi Yang[2], Mengyu Liang[3,4], Iván Higuera-Mendieta[3], Caleb Robinson[1], Anthony Ortiz[1], Min Chen[2], David B. Lobell[3,5], Inbal Becker-Reshef[1], Juan M. Lavista Ferres[1]

1 Microsoft AI for Good Lab, Redmond, WA, USA

2 Department of Forest and Wildlife Ecology, University of Wisconsin-Madison, Madison, WI, USA

3 Department of Earth System Science, Stanford University, Stanford, CA, USA

4 Department of Geographical Sciences, University of Maryland, College Park, MD, USA

5 Center on Food Security and the Environment, Stanford University, Stanford, CA, USA

***Corresponding author**: Yuchi Ma (yuchima@microsoft.com)

## Abstract

Earth observations from satellite platforms have profoundly transformed research and applications across numerous domains. However, uneven observation coverage and frequent data gaps continue to limit their ability to capture Earth's surface dynamics. Moreover, processing multi-modal Earth observations requires substantial computational resources and sophisticated data harmonization frameworks, which may not be readily accessible to many remote sensing practitioners. Earth embeddings, which are geo-encoded representations of Earth's surface generated by Geospatial Foundation models, have emerged as promising solutions to these challenges. As general-purpose representations converted from multi-source Earth observations, Earth embeddings can support a wide range of downstream tasks with no requirement of task-specific feature design and preprocessing. In this study, we first introduce the theoretical foundations of Earth embeddings. Then, we review several representative Earth embedding products. We next examine the applications of Earth embeddings across six major domains of environmental remote sensing. Finally, we systematically assess the performance of Earth embeddings in four downstream tasks, including hyperspectral image classification, individual tree crown segmentation, crop yield prediction, and aboveground biomass mapping. Our review and case studies suggest that, despite several advantages over conventional Earth observation data, current Earth embeddings do not always outperform remote sensing imagery or hand-crafted features. Earth embeddings should not be viewed as a direct replacement for conventional EO data, but rather as representations that provide stable information across space and time. Remote sensing practitioners may consider evaluating Earth embeddings' suitability for specific tasks and using them strategically, rather than assuming superior performance by default.

## 1 Introduction

Since the launch of the first Landsat satellite in 1972, Earth observations (EO) have become an indispensable resource for scientific discovery and technological advancement (Roy et al., 2014; Wulder et al., 2022). Over the past half-century, advances in satellite observation capabilities, coupled with the growing volume and accessibility of EO data, have fundamentally reshaped research and applications across numerous disciplines, including agroecosystem monitoring (Zhang et al., 2025), natural disaster management (Kumari et al., 2025), urban planning (Wellmann et al., 2020), and climate change impact assessment (Guan et al., 2023).

Despite continued improvements in the availability and quality of EO data, several challenges remain that hinder their broader application. A key challenge arises from the uneven spatial and temporal availability of EO data, resulting from variations in satellite orbital coverage, persistent cloud cover, and sampling limitations. The resulting sparse observations and data gaps can significantly compromise the effectiveness of environmental monitoring (Ma et al., 2024). These limitations are particularly consequential for time-sensitive tasks, such as wildfire detection and deforestation alerts. Even with highly accurate models, persistent clouds and relatively long revisit times can delay observations, making timely wildfire alters and near-real-time deforestation monitoring challenging. Another significant challenge arises from the need to harmonize multimodal EO datasets, as differences in sensor design, spatial resolution, spectral response functions, and revisit frequencies complicate their integration and joint analysis. Moreover, it takes massive computational resources to download and process EO, and incorporating additional modalities would require further computational resources.

Earth embeddings have emerged as a promising solution to overcome these challenges. They are compact numerical representations of EO data learned by pretrained deep learning models, which converts EO (e.g., satellite imagery and climate reanalysis data) into abstract representations of the Earth's surface (Fang et al., 2026). The generated embeddings can support a variety of downstream applications. In general, any deep learning model that accepts EO as inputs can be used to generate Earth embeddings, but traditional models are task-specific, and their embedded features are not generalizable enough to support different tasks. In practice, Earth embedding products are generated by task-agnostic geospatial foundation models (GFMs) that are typically pretrained on large-scale and diverse datasets using supervised learning, self-supervised learning (SSL), or a hybrid of the two (Lane and Karimzadeh, 2026; Xu et al., 2024). Meanwhile, GFMs can naturally and dynamically harmonize data from multiple sources. The generated Earth embeddings are seamless and capture semantic, spatial, and temporal information about the Earth's surface (Lane and Karimzadeh, 2026; Zhu et al., 2026). Although pretraining GFMs requires many resources, the precomputed Earth embeddings can be directly used to support downstream applications by fitting a linear regression model or training a simple machine learning model. As such, Earth embeddings provide an effective means to overcome observation gaps, integrate multi-modal data, and lower computational demands.

Although Earth embeddings are increasingly used in environmental remote sensing, it remains unclear to practitioners under what conditions Earth embeddings should be preferred over raw imagery or hand-crafted remote sensing features. Producers of Earth embeddings often report superior performance of embedding-based models relative to existing approaches on benchmark datasets. However, these gains do not necessarily translate to real-world applications, where Earth embeddings may not consistently outperform models based on imagery or hand-crafted features. Therefore, it is essential to understand when Earth embeddings offer meaningful advantages and when conventional remote sensing features remain the more appropriate choice. It is also worth discussing whether Earth embeddings can completely replace remote sensing features or provide complementary information.

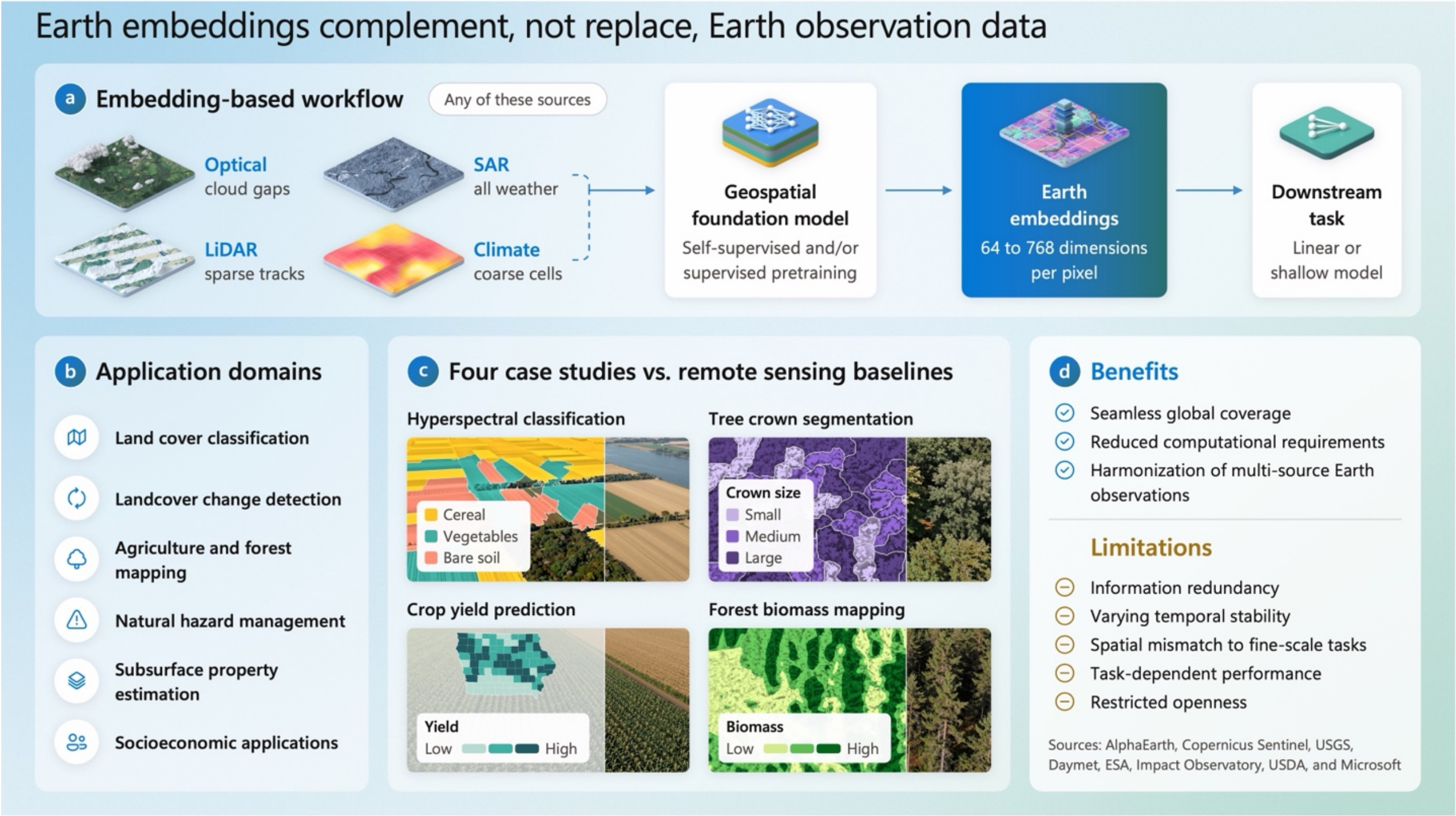

**Figure 1** This study reviews and examines Earth embeddings from a practioner's perspective, covering topics including (a) theoretical foundations of Earth embeddings and representative productions; (b) Earth embeddings' applications in six domains; (c) four case studies comparing Earth-embedding-based models with task-specific remote sensing baselines.; (d) a summary of pros, cons and practical implications of using Earth embeddings in environmental remote sensing tasks.

Existing reviews have provided comprehensive overviews of the evolution of Earth embeddings and their underlying training methodologies, primarily from the perspectives of model developers and researchers (Klemmer et al., 2026; Stewart et al., 2026). In contrast, this study examines Earth embeddings from a user- and practitioner-oriented perspective, with an emphasis on their practical utility, limitations, and appropriate use cases in environmental remote sensing (Figure 1). We conduct four case studies, along with literature reviews, that systematically compare Earth-embedding-based models with conventional remote-sensing-based approaches, with at least two Earth embedding products evaluated in each case study.

The remainder of this review is organized as follows. Section 2 introduces the theoretical foundations of GFMs and representative Earth embedding products. Section 3 reviews applications of Earth embeddings in environmental remote sensing, highlighting both successful applications and cases in which Earth embeddings provide limited or no performance improvements. Section 4 presents case studies in four environmental remote sensing areas, including hyperspectral image classification, tree crown segmentation, crop yield prediction, and forest biomass mapping. Building on the literature review and empirical comparisons, Section 5 discusses the advantages and limits of Earth embeddings and provides practical guidance on when and how Earth embeddings can be used. Finally, Section 6 summarizes the main findings and concludes the review.

# 2 Foundations of Earth Embeddings

## 2.1 Definition and Training Strategies

Earth embeddings are vector representations of areas on Earth—image patches, individual pixels, or time series—generated by models that use raw satellite imagery and other EO data (climate, elevation, gravity fields, etc.) as inputs. Here, "vector" refers to an array of numerical features learned by the model, rather than a geospatial vector geometry such as a point, line, or polygon. These embeddings are produced by GFMs that are typically trained through task-specific supervised learning and/or SSL, both involving training with large-scale datasets (Figure 2).

In task-specific supervised learning, backbones trained with ImageNet for natural image classification (Deng et al., 2009) are the original structures that provide transferable features to other computer vision tasks, and Satlas is the EO version, which is pretrained using 302M labeled satellite images under 137 categories for seven tasks (Bastani et al., 2023). However, supervised pretraining transfers only as far as its supervisory tasks are representative and diverse, and labels for environmental targets—which require field campaigns or expert interpretation—are concentrated in the regions that past programs happened to fund. Supervised features therefore inherit that footprint, and adapting a supervised backbone to a distant task generally requires unfreezing and retraining a substantial part of the network.

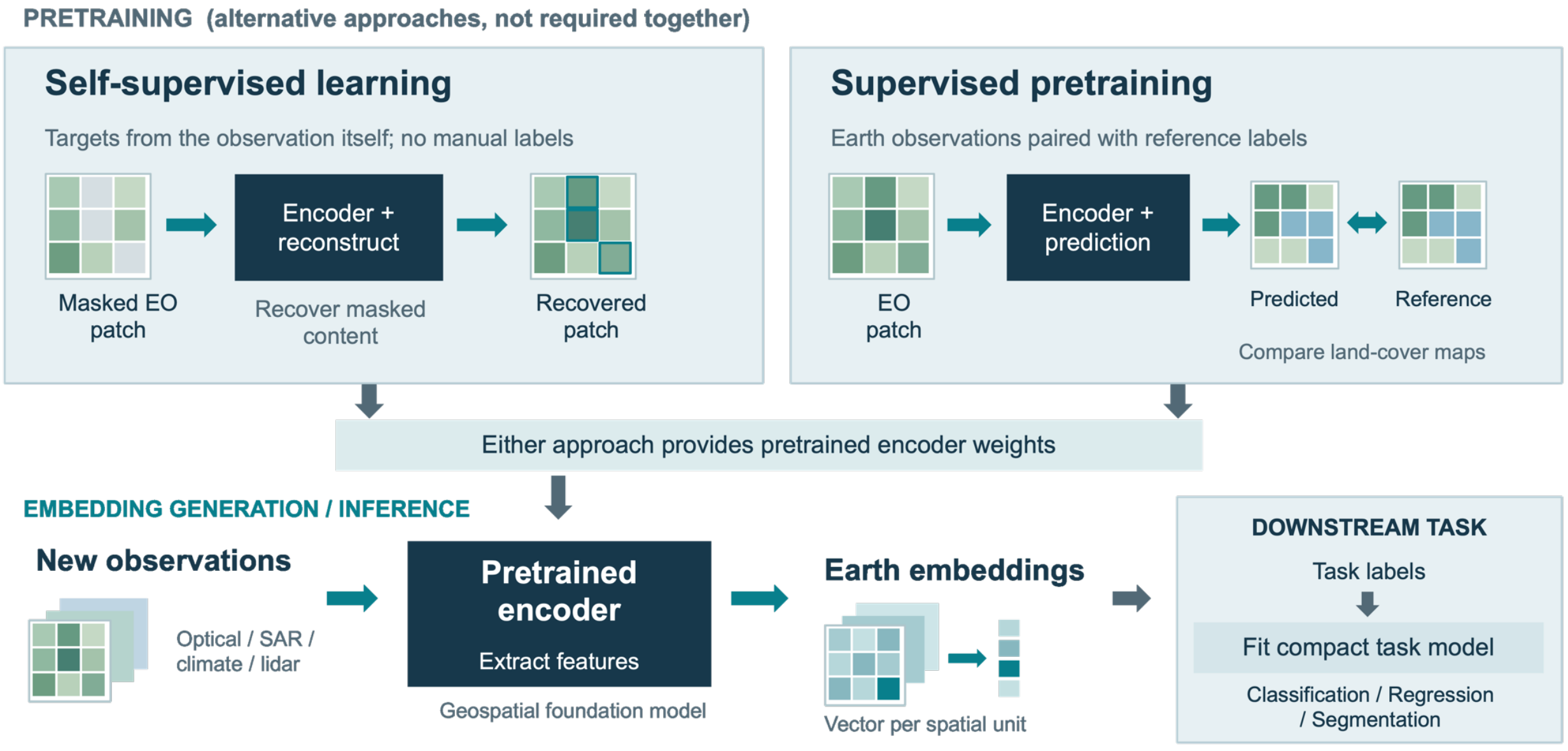

**Figure 2** The workflow of Earth embedding generation and inference. Note that masked image modeling is used as an example of self-supervised learning, while there are other self-supervised learning methods.

Unlike task-specific supervised learning, SSL requires no direct supervision. Instead, self-supervised objectives, or pretext tasks, are designed to learn general-purpose representations that transfer across multiple downstream tasks. The two main pretext tasks are *imputation*, exemplified by masked autoencoders (MAE), and *view matching*, exemplified by contrastive models. Both rest on the same assumption: that the distribution of the observed high-dimensional samples $\boldsymbol{x} \in \mathbb{R}^D$— where $\boldsymbol{x}$ may concatenate several spectral bands, time steps, or data products—has *low-dimensional support* ($d \ll D$) (Buchanan et al., 2026).

The goal of SSL is to recover a compressed representation via an encoder $f$ that maps data to features and (often) a decoder $g$ that maps the representation back to the data (Eq (1)):

$$\boldsymbol{X} \overset{f}{\rightarrow} \boldsymbol{Z} \overset{g}{\rightarrow} \widehat{\boldsymbol{X}} \tag{1}$$

Here $\boldsymbol{z} = f(\boldsymbol{x}) \in \mathbb{R}^d$ is the embedding of a single sample, and $\boldsymbol{Z} = [\boldsymbol{z}_1, \ldots, \boldsymbol{z}_n] \in \mathbb{R}^{d \times n}$ is the feature matrix, with $d \ll D$. The objective trains the pair with a reconstruction loss (Eq (2)):

$$\min_{f,g} \parallel \boldsymbol{X} - \widehat{\boldsymbol{X}} \parallel_2^2, \qquad \widehat{\boldsymbol{X}} = g(f(\boldsymbol{X})) \tag{2}$$

This is the principle behind MAE pretraining for some GFMs (Cong et al., 2022; Fuller et al., 2022). Figure 3 shows an example, in which 75% of the patches of an image $\boldsymbol{x}$ are hidden and $g \circ f$ is trained to restore it, which forces $f$ to capture the low-dimensional structure of the data. The choice of what to hide, match, or predict determines the invariance that the embedding will acquire. Masking patches teaches spatial context; masking whole bands teaches spectral structure; temporal masking teaches seasonality and change; and matching optical imagery with SAR teaches modality-invariant features (Feng et al., 2026).

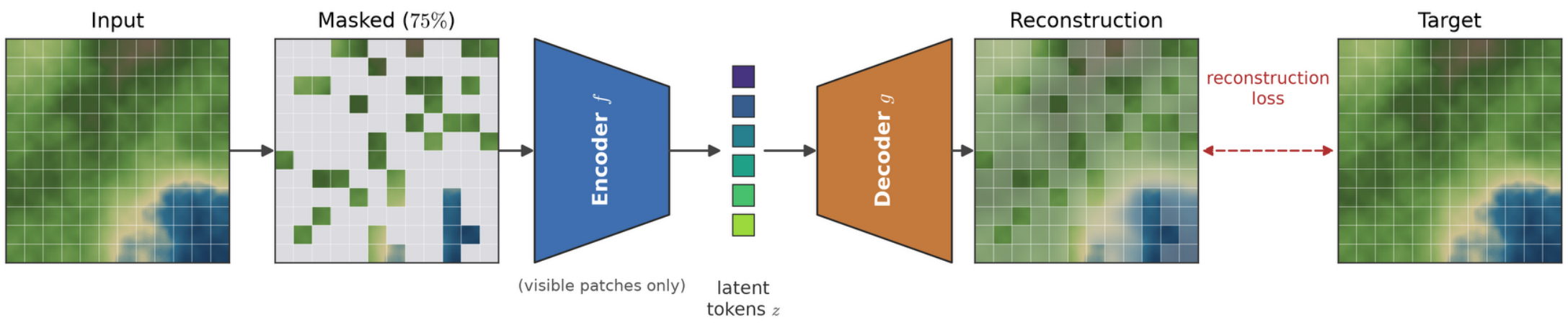

**Figure 3 Masked autoencoder (MAE).** In this pretext task the image patches are masked (75% here). The encoder $f$ embeds only the visible patches into latent tokens $\boldsymbol{z}$, and the decoder $g$ reconstructs the complete scene. The supervisory signal is the reconstruction loss between the decoded output and the original image (target).

The second main SSL strategy, contrastive view matching, is illustrated in Figure 4. For a selected view $\boldsymbol{z}$, its matching view $\boldsymbol{z}^+$ and non-matching view $\boldsymbol{z}^-$ are generated by an encoder $f$. The training objective encourages $f$ to acquire representations $\boldsymbol{z}$ that capture shared characteristics among pairs of matching views $(\boldsymbol{z}, \boldsymbol{z}^+)$ while discerning differences from non-matching samples $(\boldsymbol{z}, \boldsymbol{z}^-)$.

In EO data, matching views are typically constructed from the geometry of the data. Two acquisitions of the same location at different dates form a positive pair, teaching invariance to illumination and phenology (Yao et al., 2021); two spatially adjacent tiles form a positive pair, encoding the spatial autocorrelation of land cover (Jean et al., 2019); co-registered observations from different sensors form a cross-modal positive pair, teaching modality-invariant features (Fuller et al., 2023); and an image paired with its own coordinates forms the positive pair used by location-embedding models (Klemmer et al., 2025; Vivanco Cepeda et al., 2023). Non-matching views are, in most implementations, simply the other samples in the minibatch, drawn from different locations or under different environmental conditions.

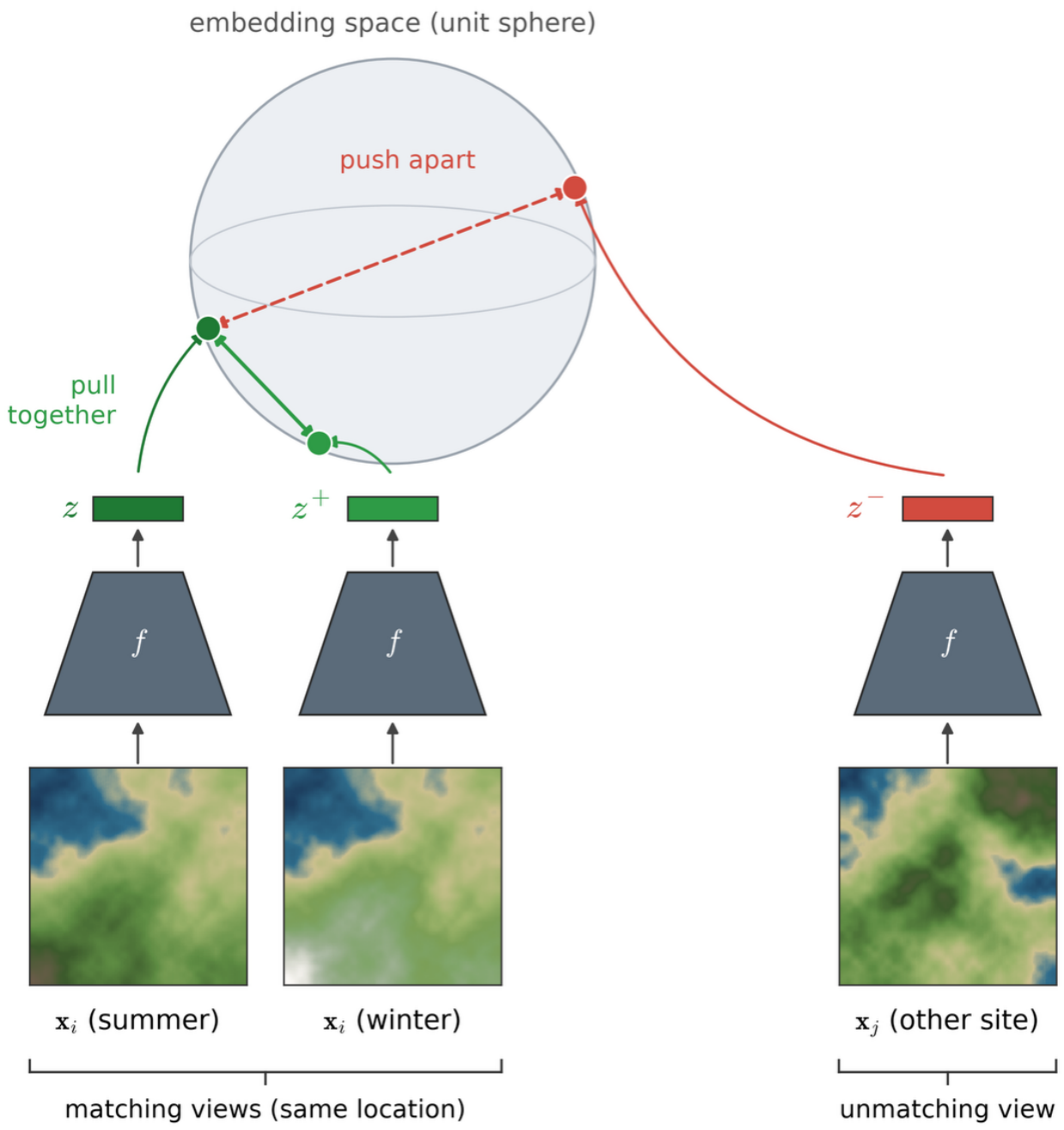

**Figure 4 Contrastive pretext.** The objective encourages similarity between a pair of matching (positive) views, $\boldsymbol{z}$ and $\boldsymbol{z}^+$, while pushing apart non-matching (negative) views $\boldsymbol{z}^-$. Here the positive pair is the same location imaged in two seasons. The model thereby learns a distance function that yields a general representation of the image content.

A third design axis concerns whether supervision enters pretraining at all. Most GFMs are SSL-pretrained and then fine-tuned with labels for a specific task, but the embedding products reviewed here are consumed frozen, so only supervision applied during pretraining affects the distributed vectors (Brown et al., 2025; Chen et al., 2026). A related but distinct strategy adds distillation from a frozen teacher (Clay Foundation, 2024), where the external signal derives from another self-supervised model rather than from labels. Joint supervision tends to improve linear separability for targets resembling the supervised heads, which matters because embedding products are typically consumed by linear probes and shallow classifiers. It does so, however, at the cost of the task-agnosticism these products claim, and it makes the composition of the training signal a practical question for users: an embedding supervised on land-cover labels cannot be evaluated on land-cover classification without circularity.

## 2.2 Representative Products

Here we describe representative Earth embedding products and the models used to generate them. We focus on products that provide a documented, user-accessible route to obtaining embeddings, whether as a precomputed global archive, an on-demand service, or a packaged pipeline that users can use locally (Table 1). GFMs that are distributed only as research checkpoints, without an accompanying embedding product, such as Galileo (Tseng et al., 2025) and TerraMind (Jakubik et al., 2025), are not reviewed here.

**AlphaEarth Foundations (AEF)**: AEF, developed by Google DeepMind, represents one of the first globally available, annual, analysis-ready Earth embedding products at 10-m spatial resolution (Brown et al., 2025). AEF integrates information from diverse data modalities, including Sentinel-2, Landsat-8/9, Sentinel-1, PALSAR-2, ERA5-Land, GEDI, GRACE, GLO-30, NLCD, and geolocated Wikipedia text. Optical and synthetic aperture radar (SAR) observations are used as model inputs, whereas the other modalities serve as reconstruction targets during training. Data from different sources are resampled to a common spatial resolution, normalized using global statistics, and processed through modality-specific input channels before being projected into a shared latent space. The latent features are subsequently concatenated and processed by a series of Space Time Precision (STP) encoder blocks designed to preserve local spatial information while modeling long-range dependencies. The resulting representations are aggregated into 64-dimensional embeddings.

**TESSERA**: TESSERA is a pixel-oriented GFM developed to address the loss of temporal information caused by conventional compositing of EO time series using SSL (Feng et al., 2026). The model employs a dual-branch encoder architecture to process annual time series of Sentinel-1 and Sentinel-2 observations and generate modality-specific representations, termed "d-pixels." For each d-pixel, two augmented views are constructed through random sampling of cloud-free observations. These views are subsequently passed through a weight-sharing dual-branch network to extract complementary representations, which are then integrated using a multilayer perceptron (MLP). The final output consists of 128-dimensional embeddings produced at a spatial resolution of 10 m with global coverage.

**OlmoEarth**: OlmoEarth is a family of fully open EO foundation models developed by the Allen Institute for AI and released across four sizes—Nano (1.4 M parameters), Tiny (6.2 M), Base (≈90 M) and Large (≈300 M)—that share a common architecture and training recipe (Herzog et al., 2026). The model is a vision transformer over multi-modal satellite image time series in which every token encodes a portion of an image together with its location, acquisition time, and sensor, allowing heterogeneous and irregularly sampled inputs to be processed as a single sequence. It is trained with a latent masked-image-modeling objective whose masking strategy and loss are designed specifically for the EO domain. OlmoEarth differs from the other products in how embeddings reach the user. Rather than publishing a fixed global archive, OlmoEarth Studio computes embeddings on demand for a user-specified area of interest, time range, encoder variant, spatial resolution, imagery source (Sentinel-2, Sentinel-1, or both) and temporal aggregation (one to twelve monthly periods), returning a cloud-optimized GeoTIFF with one signed 8-bit band per embedding dimension.

**Embedded Seamless Data (ESD)**: ESD is a global, 30 m Earth embedding database spanning 2000–2024, giving it by far the longest temporal record of any product reviewed here (Chen et al., 2026). It is built on harmonized Landsat and MODIS Terra across six spectral bands, so that the embeddings inherit a cloud-free, seam-free and sensor-bias-corrected input. The ESDNet architecture couples a spatio-temporal fusion module with a latent encoder that applies finite scalar quantization and a set of multitask heads that both generate downstream products and reconstruct the original reflectance cube. ESDNet generates lightweight Earth embeddings by compressing the annual phenological cycle into twelve monthly latent steps, yielding roughly a 340-fold reduction in data volume relative to the raw archive.

**Major TOM**: Major TOM is an open, expandable dataset framework developed by the European Space Agency (ESA)'s Φ -lab in collaboration with CloudFerro that standardizes where embeddings are computed and then publishes embeddings from several encoders on that common footprint (Czerkawski et al., 2024). Major TOM Core supplies a global, dense, gridded archive of Sentinel-1 and Sentinel-2 imagery, and the embeddings expansion attaches precomputed patch-level vectors to those same grid cells, so that outputs from different encoders are directly comparable and jointly queryable. Four global datasets anchor the release: Sentinel-2 multispectral and Sentinel-1 backscatter embedded with ResNet-50 encoders trained by DINO and MoCo respectively, and Sentinel-2 RGB embedded with the general-purpose vision models DINOv2 and SigLIP.

**Clay**: Clay is an open-source GFM designed to produce general-purpose Earth embeddings from heterogeneous sensors (Clay Foundation, 2024). It employs a ViT-based masked autoencoder (He et al., 2022) with a dynamic embedding block that generates patch tokens from the number of input bands and their central wavelengths, enabling ingestion of Sentinel-1, Sentinel-2, Landsat, MODIS, NAIP, and Land Information New Zealand (LINZ) imagery with arbitrary band counts and patch sizes. Positional encodings scaled by ground sampling distance are combined with geolocation and acquisition time, and training couples MAE reconstruction with distillation from a frozen DINOv2 teacher (Oquab et al., 2024). The frozen encoder maps a multi-band patch and its metadata to a 768-dimensional embedding for downstream fine-tuning.

**Copernicus-Embed**: Copernicus-Embed provides a global grid-level embedding map at 0.25-degree resolution, with each grid cell represented by a 768-dimensional latent vector (Wang et al., 2025). The embeddings are derived from Copernicus-FM, which is pretrained on Copernicus-Pretrain, a large-scale multi-modal EO dataset containing 18.7 million aligned observations from major Sentinel missions (Table 1). Unlike GFMs designed for fixed spectral inputs, Copernicus-FM uses dynamic hypernetworks to process both spectral and non-spectral modalities, while Fourier-encoded metadata representing geolocation, spatial coverage, and acquisition time are incorporated into the patch tokens. The model is pretrained using masked image modeling and auxiliary continual distillation, enabling it to learn representations across surface, atmospheric, and topographic observations. Copernicus-Embed is generated by aggregating Copernicus-FM-encoded multi-modal observations within each grid cell, producing compact representations with broader weather and climate contexts.

**Table 1 Specifications of representative Earth embedding products**. LS, Landsat; S1/S2/S3, Sentinel-1/-2/-3; S5P, Sentinel-5 Precursor; MODIS, Moderate Resolution Imaging Spectroradiometer; VIIRS, Visible Infrared Imaging Radiometer Suite; PALSAR-2, Phased Array type L-band Synthetic Aperture Radar-2; GEDI, Global Ecosystem Dynamics Investigation; GLO-30, Copernicus DEM Global 30 m; ERA5-Land, fifth-generation ECMWF atmospheric reanalysis over land; GRACE, Gravity Recovery and Climate Experiment; NLCD, National Land Cover Database; NAIP, National Agriculture Imagery Program; LINZ, Land Information New Zealand; DEM, digital elevation model.

| Product | Unit | Resolution | Dim | Temporal | Coverage | Source modalities | Delivery | Access |
|---|---|---|---|---|---|---|---|---|
| AEF | pixel | 10 m | 64 | annual | global, 2017–2025 | S1, S2, LS-8/9, PALSAR-2, GEDI, GLO-30, ERA5-Land, GRACE, NLCD | precomputed | Google Earth Engine |
| TESSERA | pixel | 10 m | 128 | annual | global, 2017–2025 | S1, S2 | precomputed | geotessera Python library |
| OlmoEarth | pixel | 10 / 20 / 40 / 80 m | 128 / 192 / 768 | 1–12 monthly periods | user-defined area | S1, S2 | on demand | Studio UI / API |
| ESD | pixel | 30 m | 12 monthly latent steps per year | annual | global land, 2000–2024 | LS-5/7/8/9 + MODIS Terra, 6 bands, via SDC30 | precomputed | iEarth platform |
| Major TOM | patch | 10 m | model-dependent | per acquisition | global S1/S2 footprints | S1, S2 | precomputed | Hugging Face |
| Clay | patch | sensor-dependent | 768 / 1024 | per acquisition | no global archive | S1, S2, LS, MODIS, NAIP, LINZ | run locally | Hugging Face |
| Copernicus-Embed | grid | 0.25° | 768 | static | global | S1, S2, S3, S5P, Copernicus DEM | precomputed | Hugging Face; TorchGeo GeoTIFF redistribution |

# 3 Applications

## 3.1 Land-cover Classification

Land-cover classification is among the most established applications of Earth embeddings, serving both as a practical mapping objective and as a common testbed for assessing embedding quality (Brown et al., 2025; Feng et al., 2026). Conventional pipelines often rely on engineered spectral, temporal, and contextual features. Precomputed embeddings can substantially reduce this feature-engineering burden by fitting a lightweight model on embedding vectors (Hong et al., 2024; Jakubik et al., 2023). Three broad usage patterns are emerging:

**(1) Single-product evaluation with shallow classifiers.** The dominant paradigm treats embeddings as drop-in representations for conventional remote sensing features. Brown et al. (2025) evaluated linear and $k$-nearest-neighbor classifiers on annual AEF embeddings across land-cover, crop-type and other mapping tasks, finding consistent gains over previous featurization approaches without retraining the embedding model. Khan and Ahmad (2025) demonstrated pixel- and object-based classification using AEF embeddings, illustrating the compatibility of embedding-based workflows with cloud platforms, while Murakami (2025) reported within- and cross-regional crop classification using annual embeddings in cool-climate upland agriculture.

**(2) Embedding-agnostic mapping frameworks.** A second line of work treats the embedding product as an interchangeable input layer within a standardized classification workflow. Timoner et al. (2026) calibrated classifiers in both TESSERA and AEF embeddings for environmental-descriptor classification and used the resulting descriptors to construct land-cover maps in Switzerland, Brazil, and Spain. Such workflows decouple downstream mapping from any single embedding product and could therefore facilitate operational applications as embedding products evolve.

**(3) Comparative benchmarking against conventional features.** A third strand asks when embeddings outperform well-tuned conventional inputs. Systematic benchmarks show that foundation-model representations do not uniformly outperform supervised baselines, with gains that are task-dependent and sensitive to sensor modality and evaluation setting (Marsocci et al., 2024).

Overall, the evidence base for embedding-driven land-cover classification is growing rapidly but remains uneven (Gilch et al., 2026; Marsocci et al., 2024). Several open questions remain. It is unclear whether fine-grained thematic classes are better supported by pixel-level embedding fields (e.g., AEF and TESSERA) or by patch-level representations (e.g., Clay). The consistency of embedding-based maps across successive annual product releases has yet to be examined. It also remains uncertain to what extent fixed, task-agnostic embeddings limit the discrimination of spectrally or temporally subtle classes.

## 3.2 Land-cover Change Detection

Although most GFMs were not originally developed for change detection, their temporally consistent embedding spaces have increasingly motivated methodological developments in land-cover change detection (Li et al., 2024; Yu et al., 2024). Three complementary methodological paradigms are emerging (Ding et al., 2025; Peng et al., 2025):

**(1) Temporal Embedding Comparison.** When embeddings are generated for the same location at two dates, land-cover change appears as a measurable shift in vector space (Gilch et al., 2026; Liu et al., 2025). Li et al. (2025) projected Harmonized Landsat Sentinel-2 time series into a low-dimensional embedding space and incorporated them into a change detection framework. Their results showed that compact embeddings detected land disturbances at accuracy levels comparable to the full multispectral stack, while reducing storage by an order of magnitude. Heiman (2026) applied a cosine similarity threshold to annual AEF embeddings to track reforestation sites in Brazil's Atlantic Forest from 2017 to 2024. Cross-referencing vectorized change polygons with MapBiomas products showed spatial correspondence with documented deforestation and urban expansion patterns. However, cosine similarity alone cannot reveal the direction of a transition (e.g., forest to cropland vs. cropland to forest), underscoring the need for complementary classification.

**(2) Supervised Detection and Model Adaptation.** When change labels are available, embeddings can serve as input features for supervised change classification. Brown et al. (2025) showed that, when trained with change labels, AEF outperformed competing representations. An alternative strategy adapts pretrained models directly to the bi-temporal task. The Bi-Temporal Adapter Network (Li et al., 2024) attaches trainable bi-temporal adapters to a frozen foundation model (e.g. CLIP), while change-aware self-supervised pretraining (Feng et al., 2024) learns representations of changed regions from unlabeled image pairs before fine-tuning. Both achieve strong performance but require additional training and labeled data, in contrast to embedding-based approaches that reuse precomputed representations with only a lightweight classifier.

**(3) Unsupervised Change Monitoring.** Zhang et al. (2022) adapted the SimCLR contrastive learning framework specifically for wildfire burn scar mapping, using cosine distance between pre-fire and post-fire encoded images as the change signal, and outperformed conventional baselines without any labeled data. Albughdadi et al. (2025) measured embedding divergence between pre-event and post-event Sentinel-1 and Sentinel-2 acquisitions processed through Clay, detecting disaster-driven changes without task-specific training. These unsupervised approaches are particularly valuable for rapid assessment following natural disasters, where labeled data are unavailable.

Embedding-based change detection remains an emerging research direction. Most of GFMs cited as enabling tools in the studies above were designed as general-purpose representation frameworks rather than change detection methods. Their growing adoption in monitoring workflows reflects the practical utility of temporally consistent embeddings, but dedicated benchmarks and peer-reviewed methodological studies tailored to land-cover change detection remain relatively limited (Yang et al., 2024). Expanding this evidence base represents a potential direction for future research.

## 3.3 Agricultural and Forest Monitoring

Agricultural and forest monitoring requires frequent and cloud-free observations, but infrequent EO and cloud-cover-caused data gaps can significantly reduce the effectiveness of satellite-based monitoring in agricultural and forest landscapes. Emerging Earth embedding products offer a promising pathway to mitigate these limitations by filling data gaps in conventional EO and improving the spatial and temporal continuity.

**(1) Crop type mapping.** Crop type mapping provides fundamental information about what crops are grown, where they are grown, and how agricultural landscapes change over time. It is adopted as one of the standard downstream tasks to evaluate the effectiveness of Earth embeddings. Google DeepMind (Brown et al., 2025) evaluated AEF embeddings in three crop type mapping tasks, including binary cropland classification in Africa (Kerner et al., 2024), multi-level crop classification in Canada (Fisette et al., 2013), and crop type classification in Ethiopian smallholder wheat-based farming systems (Blasch et al., 2024). Their internal evaluations showed that AEF-based models significantly reduced classification errors compared to the second-best models in all tasks. Similarly, a small neural network classifier was trained using TESSERA embeddings (Feng et al., 2026) for crop type classification and evaluated using the INVEKOS Austrian crop dataset in comparison with three alternatives: a random forest classifier trained on engineered satellite features, the PRESTO foundation model, and the AEF embeddings. The TESSERA-based model consistently outperformed baselines. Although promising, Earth-embedding-based crop type mapping models do not always outperform RS-based methods. For example, CropNet, a CNN-based crop type classifier based on time-series Sentinel-2, outperformed AEF-based models by more than 20% and 40% in cross-continent and cross-hemisphere crop type mapping experiments, respectively (Tong and Wang, 2025).

**(2) Crop yield mapping.** Crop yield mapping enables stakeholders to anticipate agricultural production before harvest (in-season yield prediction) (Ma et al., 2021) and understand local trends and variability of yields (post-season yield mapping) (Ma et al., 2024). Ma et al. (2026) evaluated Earth-embedding-based crop yield estimation for corn, soybeans, and winter wheat in the U.S. AEF-based models slightly outperformed RS-based models in local experiments (training and testing in the same region), but have issues with transferability, interpretability, and stability.

**(3) Forest Aboveground Biomass (AGB) Mapping.** Recent studies have tested AEF embeddings for biomass mapping using forest inventory data across diverse forest types, including ancient deciduous forest remnants in Iran, mixed coniferous-deciduous plantations in eastern China, and mountainous mixed-species forests in Spain (Jin et al., 2026; Pascual and Guerra-Hernández, 2026; Sohrabi, 2026). These studies generally estimate biomass for a single year by training models that link field-derived AGB to AEF embeddings. The reported results suggest that, with representative training data, Earth embeddings can support spatial biomass prediction and gap filling. However, forest biomass monitoring also requires models that are temporally reliable. To date, few studies have tested whether embedding-based models can detect and quantify biomass change from abrupt or gradual processes such as deforestation, degradation, and recovery. One exception is Ball et al. (2026), who evaluated AEF and TESSERA embeddings for tree species mapping across two years and found greater degradation in temporal performance for AEF. Whether similar limitations affect biomass change estimation needs future testing.

**(4) Agricultural practice mapping.** Agricultural practices, such as tillage, cover crop, irrigation, and crop rotation, directly influence crop productivity. Agricultural practice mapping reveals how crops are managed, which is important to cropland productivity and environmental sustainability (Lobell et al., 2025). Ma et al. (2026) presented a systematic assessment of AEF embeddings for tillage type classification and cover crop detection in the U.S., finding that AEF-based models did not necessarily outperform traditional RS-based models. Yang et al. (2026b) assessed AEF embeddings for irrigated cropland mapping and found that AEF outperformed conventional optical and SAR-based features. AEF was also found to have stable cross-year transferability but limited spatial generalizability.

## 3.4 Natural Hazard Management

Hazard applications invert the premise that makes annual embeddings attractive elsewhere. Land-cover and biomass mapping benefit from temporal aggregation, whereas fire, flood and landslide events are short-lived and their timing is the quantity of interest. The literature divides according to whether the target is a persistent post-event signature, a static susceptibility surface, or a rapid damage assessment, although the third of these remains largely out of reach.

**(1) Wildfire.** Seydi (2025) trained bi-temporal Siamese networks on AEF embeddings using Monitoring Trends in Burn Severity (MTBS) reference data in the conterminous United States and evaluated transfer to seventeen regions in Europe, reporting 95% overall accuracy, with particular strength in delineating fire boundaries and partially burned vegetation. The author identifies the governing constraint precisely: the annual embedding bands encode seasonal and temporal variation but are not assigned to specific months, which limits their use where intra-annual timing matters. Another study of fire severity used AEF to estimate the satellite-derived fire severity index, achieving $R^2$ scores of 0.29–0.87 across global biomes (Mahato et al., 2026). They suggested that Earth embeddings are promising to complement the conventional index and overcome challenges such as smoke and cloud interference, canopy obstruction in dense forest, and gaps between fire occurrence and satellite revisit.

**(2) Landslides.** Landslide susceptibility mapping conventionally depends on landslide conditioning factors (LCFs) whose availability, heterogeneity, and preprocessing uncertainty constrain reliability. Cheng et al. (2026) evaluated AEF embeddings as a substitute, comparing retained principal components and the full 64 bands against conventional conditioning factors across Nantou County (Taiwan), Hong Kong and part of Emilia-Romagna (Italy), finding Earth-embedding-based models outperform LCFs across all regions. A complementary study used embeddings to construct annual landslide inventories along the middle and lower Nujiang River (China) from 2017 to 2025, recovering strong spatial clustering, marked interannual variability, and persistent hotspots (Liu et al., 2026).

**(3) Floods.** Flood work has concentrated on damage prediction rather than inundation mapping. Kawakami et al. (2026) incorporated AEF embeddings as conditioning inputs in a CNN-based continental-scale daily pluvial flood damage model alongside terrain attributes. The proposed system outperformed three tree-based baselines and captured the high-cost claims that baselines underpredicted. AEF helps the model adapt the rainfall-response relationship to local geography and built environment. When the model is conditioned on AEF only, performance is close to the full model and close to terrain-only conditioning, which suggests AEF captures much of the local hydrologic context that hand-engineered terrain descriptors capture, without explicit hydrological feature engineering.

Across applications in all three hazards, annual products support pre-event risk characterization well but post-event rapid assessment poorly. This is the clearest domain in which the temporal resolution of current embedding products, rather than their representational quality, determines what is feasible.

## 3.5 Soil, Water and Subsurface Property Estimation

Soil, hydrological and subsurface targets share a property that sets them apart from the applications reviewed above: the quantity of interest is never directly observed. Soil organic carbon lies beneath a vegetation canopy for most of the year, water-table depth lies beneath the surface entirely, and runoff integrates processes across a whole basin. Conventional workflows therefore depend on carefully assembled proxy covariates, and it is precisely this covariate assembly that Earth embeddings promise to replace. Three strands of work have emerged.

**(1) Soil properties.** Digital soil mapping is constrained by the scarcity of bare-soil observation windows. A comparison of bare-soil multispectral composites, vegetation indices, AEF embeddings and a hybrid geostatistical–ML model across croplands in Italy, France, and Taiwan found that models trained on embedding covariates consistently outperformed the alternatives, suggesting that soil organic carbon assessment need not depend on bare-soil imagery (Castaldi et al., 2026). The authors caution that correlations between embedding covariates and soil organic carbon differed across the three datasets, which may limit transfer. Embeddings have also been shown to serve as effective covariates for soil organic carbon and water-table depth in peatlands, performing comparably to expert-derived remote sensing inputs with substantially less preparation effort (Koch et al., 2026).

**(2) Wetlands and surface water.** Ryan et al. (2026) compared a conventional multi-sensor workflow built from 29 Sentinel-1 GRD scenes and 294 cloud-free Sentinel-2 Level-2A images against a cluster-guided random forest on AEF embeddings for wetland vegetation mapping on the Narran Lake floodplain in Australia. Both approaches reached comparable accuracy (overall accuracy 0.985–0.991, Cohen's $\kappa$ 0.977–0.990), but the embedding-based maps showed markedly better spatial coherence, with lower edge density, local entropy and patch fragmentation. Discrepancies concentrated in shrubland and herbaceous wetland classes whose spectral and structural signatures vary with hydrological and disturbance regime. Spectrally homogeneous classes such as open water and snow require comparatively few embedding dimensions to characterize (Benavides-Martinez et al., 2026), so the marginal gain over a simple water index is likely small for permanent open water and largest in ambiguous vegetated transition zones.

**(3) Hydrological and subsurface characterization.** Beyond the surface, embeddings have been used to predict runoff across time and among river basins (Ou and Zheng, 2026), and to map subsurface properties (Nakata et al., 2026). Across the applications, AEF-informed models outperformed traditional baselines that rely only on static basin attributes. These applications are notable because the target variable is not directly observable from above, which makes them a useful test of how much of an embedding's predictive power derives from correlated surface expression rather than from the target itself.

## 3.6 Socioeconomic Applications

**(1) Urban form and local climate.** Ko Ko and Atzberger (2026) mapped fine-scale local climate zones across five Swiss cities with an attention-based U-Net, comparing TESSERA and AEF embeddings against Sentinel-1/2 composites while testing multi-city transferability. TESSERA consistently outperformed both alternatives, although year-to-year transfer remained an open problem. Gong et al. (2026) also benchmarked three Earth embedding families (AlphaEarth, Prithvi, and Clay) for urban signal prediction across six U.S. metropolitan areas. They found that Earth embeddings captured substantial urban variation, with the highest predictive skill for outcomes more directly tied to built-environment structure (e.g., chronic health burdens and dominant commuting modes), while indicators relevant to fine-scale behavior and local policy remain difficult to infer.

**(2) Wealth, poverty, and population.** Pettersson and Daoud (2025) predicted cluster-level wealth indices across sub-Saharan Africa from AEF embeddings using a graph neural network with a probabilistic loss designed to accommodate the deliberate coordinate displacement in Demographic and Health Survey data, evaluated on 37 surveys from 2017–2023. Notably, the embedding representation of their study area occupies 45.6 MB against 51.5 GB for an equivalent Sentinel-2 dataset. A two-stage approach has also been applied to improve the spatial resolution of poverty estimates in Mexico, benchmarking AEF-based models against DINOv3 variants (Salas et al., 2026). In another study (Hou et al., 2026), AEF-based population models have been used to generate annual maps at 1 km globally and 100 m for eighteen cities, positioned as complementing temporal gaps in existing gridded population products, which achieved median $R^2$ of 0.82–0.96 under random-split validation and fell by 0.46–0.67 under spatial block cross-validation.

# 4 Case Studies

Most existing applications focus on using one Earth embedding (mostly AEF) for one task (classification, segmentation, or regression). A lack of standardized training and evaluation protocols has led to disagreements across studies of Earth embedding. To provide a comprehensive evaluation of Earth embeddings for practitioners with different application requirements, this section conducts case studies in four environmental remote sensing domains in the category of classification (land-cover classification), segmentation (tree crown segmentation), and regression (crop yield prediction and above-ground biomass estimation). Two or more Earth embeddings are benchmarked in each case study and compared with task-specific remote sensing predictors. Within each case study, the training and evaluation protocols are held identical across all compared models, ensuring a fair comparison.

## 4.1 Hyperspectral Image (HSI) Classification

In this case study, we adopted three representative embeddings, including AEF, Clay, and TESSERA, and explored whether they could improve hyperspectral image (HSI) classification performance on the Augsburg dataset (Hong et al., 2021). The Augsburg scene covers the city of Augsburg in Bavaria, southern Germany, and the HSI retains 180 bands across 0.4–2.5 $\mu$m, covering seven land-cover classes (Hu et al., 2023). Meanwhile, we queried the 2018 collection of AEF and TESSERA over each scene and resampled them to the 30 m HSI grid. For Clay, we formed a 2018 Sentinel-2 L2A median composite, ran the Clay large encoder to obtain 1024-dimensional tokens, reduced them to 64 channels via PCA, and resampled to the same grid (Clay Foundation, 2024). We employed two sampling strategies to provide a comprehensive evaluation. The random fifty-pixel-per-class strategy drew fifty samples from every class of the full ground truth and left the remaining annotated pixels as testing samples. The spatially disjoint strategy used two disjoint masks that placed the training and testing pixels in geographically separated regions of the same scene (Yang et al., 2023).

A popular two-branch model, transformer-based heterogeneously salient graph representation (THSGR) network (Yang et al., 2026a), was adopted as a backbone. Each embedding was evaluated both as a replacement input and in combination with HSI via the second branch for joint classification, while the HSI baseline used the hyperspectral branch alone. We assessed classification performance using the overall accuracy (OA), the average accuracy (AA), and the kappa coefficient (kappa). Figure 5 shows the input HSI false color together with each embedding product resampled to the HSI grid. Figure 6 illustrates classification maps under both the random and the disjoint sampling strategies, and Table 2 reports the corresponding OA, AA, and kappa.

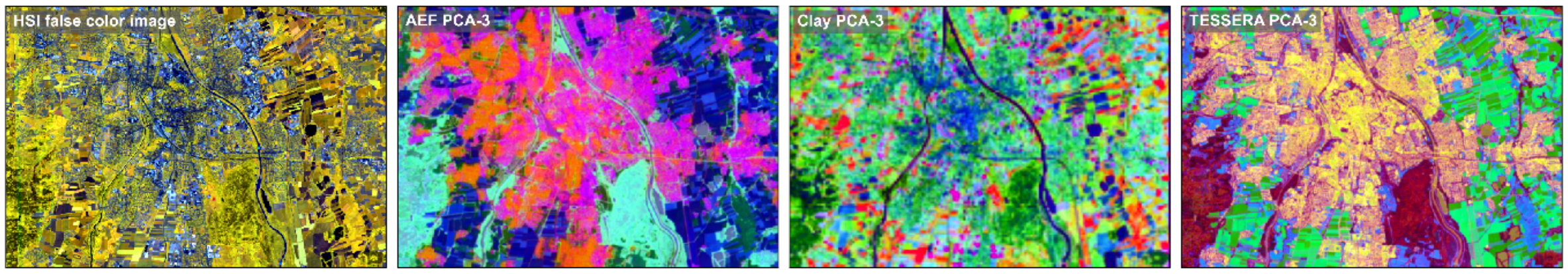

**Figure 5** Visualization for HSI, AEF, Clay, and TESSERA embeddings on the Augsburg data.

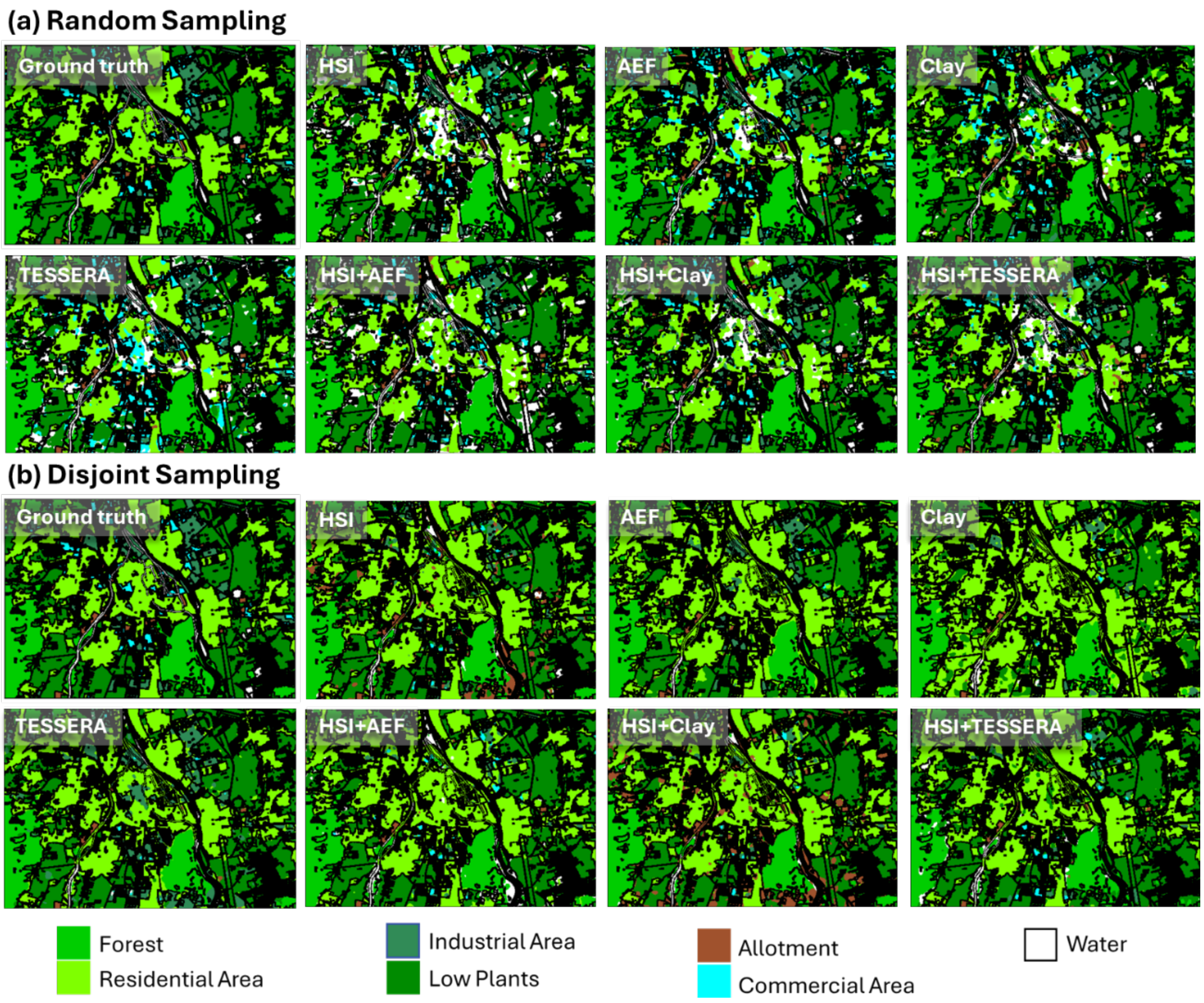

**Figure 6** Classification maps for different approaches on the Augsburg data. (a) Random sampling. (b) Disjoint sampling.

**Table 2** Augsburg test-set accuracy (%), reported as mean ± standard deviation. Within each sampling strategy, the best result is in **bold** and the second best is <u>underlined</u>.

| Approach | Sampling strategy | OA (%) | AA (%) | Kappa (%) |
|---|---|---|---|---|
| HSI | random | **85.67 ± 2.91** | <u>83.71 ± 1.00</u> | **80.49 ± 3.61** |
| AEF | | 85.00 ± 5.41 | **84.50 ± 2.14** | 79.65 ± 6.74 |
| Clay | | 81.36 ± 2.74 | 80.87 ± 0.69 | 74.56 ± 3.19 |
| TESSERA | | 82.61 ± 4.09 | 83.25 ± 2.25 | 76.66 ± 5.05 |
| HSI + AEF | | 84.83 ± 1.12 | 83.21 ± 1.36 | 79.43 ± 1.42 |
| HSI + Clay | | 84.98 ± 4.54 | 82.92 ± 2.46 | 79.67 ± 5.65 |
| HSI + TESSERA | | <u>85.16 ± 3.83</u> | 82.84 ± 1.50 | <u>79.88 ± 4.76</u> |
| HSI | disjoint | <u>87.58 ± 2.78</u> | **57.61 ± 3.49** | **82.05 ± 3.78** |
| AEF | | 86.07 ± 3.17 | 56.54 ± 3.99 | 79.46 ± 4.89 |
| Clay | | 74.76 ± 4.52 | 45.93 ± 2.89 | 62.16 ± 7.15 |
| TESSERA | | 86.80 ± 1.91 | 57.46 ± 1.97 | 80.64 ± 2.86 |
| HSI + AEF | | **87.60 ± 1.00** | 56.59 ± 2.59 | 81.97 ± 1.31 |
| HSI + Clay | | 85.47 ± 2.82 | 53.87 ± 4.82 | 78.88 ± 4.12 |
| HSI + TESSERA | | 85.88 ± 1.46 | 55.22 ± 3.88 | 79.65 ± 1.86 |

On the Augsburg scene, all approaches except Clay reached similar accuracy, and most differences among them were smaller than the run-to-run standard deviation (Table 2). Under random sampling, the HSI baseline obtained the highest mean OA (85.67%) and kappa (80.49%). Used as replacements, the Earth embeddings performed close to HSI: AEF reached 85.00% OA and the highest AA (84.50%), TESSERA reached 82.61% OA, and Clay was the lowest at 81.36% OA. Combining the embeddings with HSI did not improve on the baseline, as the three fusion variants reached 84.83–85.16% OA. Under disjoint sampling, OA remained high for all approaches except Clay (85.47–87.60%), but AA dropped to 46–58%. Per-class accuracies showed that every approach recognized the dominant forest and residential classes well (83–99%), as did all approaches except Clay for low plants (87–93% vs. 58%), whereas commercial area was almost never recognized in the spatially separated test region (below 5% for all approaches) and water was recognized in at most 36% of its test pixels. HSI + AEF (87.60%) and HSI (87.58%) achieved the highest OA, AEF and TESSERA used alone reached 86.07% and 86.80%, and Clay again performed worst (74.76% OA and 62.16% kappa).

The classification maps in Figure 6 were consistent with these quantitative results. Under both sampling strategies, all approaches recovered the overall land-cover structure of the scene, and the HSI, embedding-only, and fusion maps were broadly similar. Under random sampling, most maps contained spurious water labels in the dense city center, and the TESSERA map additionally contained commercial-area patches there. The Clay maps were smoother, with simplified class boundaries, consistent with its coarser 80 m token footprint. Under disjoint sampling, all maps were dominated by forest, residential area, and low plants, and commercial area was largely absent, in line with the low AA; the HSI + Clay map also contained spurious allotment patches.

These results suggest that, at the 30 m grid of this scene, the Earth embeddings already capture most of the class-discriminative information needed to separate these broad urban and vegetation classes. AEF and TESSERA alone separated the seven classes almost as well as the 180-band HSI, presumably because the multi-sensor, multi-temporal observations compressed into these embeddings characterize land use through its spectral and phenological behavior. Clay was consistently the weakest, plausibly because its 80 m tokens blur the boundaries of small urban parcels when resampled to 30 m, and because it was derived from a single annual median composite.

## 4.2 Individual Tree Crown Segmentation

Here, we compared high-resolution aerial imagery with Earth embedding features for individual tree crown segmentation. We used four-band imagery from the National Agriculture Imagery Program (NAIP), including Red, Green, Blue, and Near Infrared (NIR), at 0.6 m ground sample distance (U.S. Department of Agriculture, 2024). We assembled 103 patches of $256 \times 256$ pixels across the conterminous United States, each paired with individual crown polygons and a crown-boundary layer that were manually delineated and cross-checked by experts. The patches span a range of canopy conditions, from relatively open landscapes to dense forest, providing variation in both crown size and canopy density. The dataset contains 12,130 crowns in total, with 8,360 crowns in the training set and 3,770 in the test set. The two splits had similar distributions of canopy cover, crown size, and crown count, reducing the potential influence of differences in target composition between training and test data.

We evaluated AEF, Clay, and TESSERA. AEF was used at its native 64 dimensions, whereas Clay and TESSERA were reduced to 64 channels via PCA. All embeddings were then resampled to the 256 × 256 NAIP grid. AEF and TESSERA have a native resolution of 10 m, whereas Clay was generated from NAIP with an effective token footprint of approximately 4.8 m. Using U-Net as the segmentation backbone (Ronneberger et al. 2015), we compared the NAIP baseline with the three embedding-only, which used each embedding as the replacement input, and three embedding-augmented models that concatenated each embedding with the NAIP bands. Performance was evaluated using F1, recall, precision, mIoU, and area ratio (Table 3), with additional analysis by reference crown size (Table 4).

**Table 3** Individual tree crown segmentation results for different approaches. Higher is better for the first four columns; the area ratio is the total predicted crown area divided by the total annotated crown area, so 1.00 is ideal. The best result is in **bold** and the second best is <u>underlined</u>.

| Approach | F1 | Recall | Precision | mIoU | Area ratio |
|---|---|---|---|---|---|
| NAIP | $0.74 \pm 0.01$ | $0.78 \pm 0.02$ | $0.70 \pm 0.01$ | $\mathbf{0.72 \pm 0.01}$ | $1.10 \pm 0.04$ |
| AEF | 0.63±0.01 | 0.71±0.04 | 0.57±0.03 | 0.63±0.01 | 1.25±0.13 |
| Clay | 0.67±0.01 | 0.65±0.02 | 0.71±0.01 | 0.68±0.01 | 0.91±0.04 |
| TESSERA | 0.57±0.03 | 0.68±0.07 | 0.50±0.01 | 0.58±0.02 | 1.35±0.12 |
| NAIP + AEF | <u>$0.72 \pm 0.01$</u> | $0.69 \pm 0.04$ | $\mathbf{0.76 \pm 0.02}$ | <u>$0.72 \pm 0.01$</u> | $\mathbf{0.91 \pm 0.06}$ |
| NAIP + Clay | $0.72 \pm 0.01$ | $0.70 \pm 0.04$ | <u>$0.74 \pm 0.01$</u> | $0.71 \pm 0.01$ | <u>$0.94 \pm 0.07$</u> |
| NAIP + TESSERA | $0.68 \pm 0.04$ | $0.69 \pm 0.07$ | $0.67 \pm 0.02$ | $0.68 \pm 0.03$ | $1.04 \pm 0.08$ |

**Table 4** Individual tree crown segmentation results for different crown diameters. Crowns are binned by equivalent-circle diameter; the header gives the number of reference crowns in each bin and the bin's share of the total annotated crown area. Best in each column in **bold**, parentheses give 95% confidence intervals.

| Crown diameter | $< 3$ m | 3–5 m | 5–8 m | 8–15 m | $\geq 15$ m |
|---|---|---|---|---|---|
| Ground truth crowns | 419 | 956 | 1,263 | 1,037 | 95 |
| Share of crown area | 1.0% | 7.4% | 24.4% | 54.1% | 13.1% |
| | | | *Recall* | | |
| NAIP | **0.57** (0.33–0.76) | **0.66** (0.49–0.78) | **0.73** (0.63–0.81) | **0.81** (0.75–0.85) | **0.82** (0.77–0.87) |
| AEF | 0.41 (0.11–0.73) | 0.55 (0.24–0.78) | 0.66 (0.41–0.83) | 0.75 (0.53–0.91) | 0.74 (0.55–0.89) |
| Clay | 0.22 (0.09–0.37) | 0.39 (0.19–0.54) | 0.56 (0.36–0.71) | 0.70 (0.51–0.82) | 0.78 (0.61–0.88) |
| TESSERA | 0.40 (0.12–0.72) | 0.51 (0.23–0.75) | 0.61 (0.41–0.78) | 0.72 (0.55–0.85) | 0.75 (0.62–0.87) |
| NAIP + AEF | 0.54 (0.34–0.73) | 0.65 (0.48–0.78) | 0.67 (0.51–0.79) | 0.70 (0.50–0.82) | 0.74 (0.59–0.84) |
| NAIP + Clay | 0.44 (0.20–0.67) | 0.55 (0.30–0.73) | 0.63 (0.43–0.79) | 0.73 (0.56–0.84) | 0.79 (0.67–0.87) |
| NAIP + TESSERA | 0.49 (0.29–0.71) | 0.59 (0.41–0.74) | 0.65 (0.52–0.75) | 0.73 (0.61–0.81) | 0.73 (0.63–0.83) |
| *Precision* | | | | | |
| NAIP | 0.46 (0.36–0.56) | 0.69 (0.61–0.76) | **0.83** (0.79–0.86) | **0.90** (0.87–0.92) | **0.92** (0.86–0.96) |
| AEF | 0.30 (0.27–0.32) | 0.45 (0.42–0.50) | 0.59 (0.55–0.62) | 0.70 (0.66–0.73) | 0.75 (0.68–0.81) |
| Clay | 0.34 (0.25–0.39) | 0.57 (0.52–0.62) | 0.74 (0.71–0.77) | 0.84 (0.82–0.86) | 0.89 (0.87–0.91) |
| TESSERA | 0.28 (0.22–0.30) | 0.43 (0.38–0.49) | 0.58 (0.54–0.62) | 0.69 (0.65–0.73) | 0.74 (0.67–0.80) |
| NAIP + AEF | 0.50 (0.37–0.59) | 0.70 (0.63–0.77) | **0.83** (0.79–0.86) | 0.89 (0.86–0.92) | 0.90 (0.85–0.94) |
| NAIP + Clay | **0.51** (0.42–0.57) | **0.71** (0.66–0.76) | 0.82 (0.79–0.85) | 0.89 (0.86–0.91) | **0.92** (0.89–0.94) |
| NAIP + TESSERA | 0.49 (0.38–0.56) | 0.68 (0.61–0.75) | 0.82 (0.78–0.86) | 0.89 (0.85–0.92) | 0.90 (0.83–0.95) |
| *F1* | | | | | |
| NAIP | **0.51** (0.36–0.62) | **0.68** (0.56–0.75) | **0.78** (0.71–0.82) | **0.85** (0.82–0.88) | **0.87** (0.84–0.89) |
| AEF | 0.34 (0.16–0.42) | 0.50 (0.32–0.58) | 0.62 (0.47–0.69) | 0.72 (0.59–0.80) | 0.74 (0.62–0.83) |
| Clay | 0.27 (0.14–0.36) | 0.46 (0.28–0.56) | 0.64 (0.48–0.72) | 0.76 (0.63–0.83) | 0.83 (0.73–0.89) |
| TESSERA | 0.33 (0.15–0.41) | 0.46 (0.29–0.54) | 0.59 (0.47–0.66) | 0.70 (0.60–0.76) | 0.74 (0.66–0.81) |
| NAIP + AEF | **0.51** (0.37–0.63) | 0.67 (0.56–0.74) | 0.74 (0.63–0.81) | 0.79 (0.64–0.86) | 0.81 (0.71–0.88) |
| NAIP + Clay | 0.47 (0.27–0.60) | 0.61 (0.42–0.72) | 0.71 (0.56–0.80) | 0.80 (0.69–0.86) | 0.85 (0.77–0.89) |
| NAIP + TESSERA | 0.49 (0.34–0.60) | 0.63 (0.50–0.71) | 0.73 (0.63–0.79) | 0.80 (0.71–0.85) | 0.81 (0.74–0.86) |
| *IoU* | | | | | |
| NAIP | 0.34 (0.22–0.45) | **0.51** (0.39–0.60) | **0.64** (0.55–0.70) | **0.74** (0.69–0.78) | **0.76** (0.72–0.81) |
| AEF | 0.21 (0.09–0.27) | 0.33 (0.19–0.41) | 0.45 (0.31–0.53) | 0.57 (0.42–0.66) | 0.59 (0.45–0.72) |
| Clay | 0.15 (0.07–0.22) | 0.30 (0.17–0.39) | 0.47 (0.32–0.57) | 0.62 (0.46–0.71) | 0.72 (0.57–0.80) |
| TESSERA | 0.20 (0.08–0.26) | 0.30 (0.17–0.37) | 0.42 (0.31–0.50) | 0.54 (0.43–0.62) | 0.59 (0.50–0.68) |
| NAIP + AEF | **0.35** (0.23–0.45) | **0.51** (0.39–0.59) | 0.59 (0.46–0.68) | 0.65 (0.47–0.75) | 0.68 (0.55–0.78) |
| NAIP + Clay | 0.31 (0.16–0.42) | 0.44 (0.26–0.56) | 0.55 (0.39–0.67) | 0.67 (0.53–0.76) | 0.74 (0.63–0.81) |
| NAIP + TESSERA | 0.32 (0.21–0.43) | 0.46 (0.34–0.55) | 0.57 (0.46–0.65) | 0.66 (0.56–0.73) | 0.68 (0.60–0.75) |
| *Area ratio* | | | | | |
| NAIP | 1.24 (0.76–1.67) | **0.96** (0.71–1.13) | 0.88 (0.77–0.98) | 0.90 (0.83–0.96) | 0.89 (0.82–0.99) |
| AEF | 1.38 (0.38–2.46) | 1.21 (0.53–1.75) | 1.12 (0.70–1.41) | 1.08 (0.76–1.30) | <u>0.98</u> (0.75–1.18) |
| Clay | 0.66 (0.32–1.03) | 0.68 (0.35–0.95) | 0.75 (0.48–0.96) | <u>0.83</u> (0.60–0.98) | 0.88 (0.68–1.00) |
| TESSERA | 1.42 (0.45–2.56) | 1.19 (0.54–1.81) | **1.05** (0.70–1.36) | **1.03** (0.79–1.24) | **1.01** (0.83–1.19) |
| NAIP + AEF | 1.10 (0.67–1.48) | 0.93 (0.68–1.12) | 0.81 (0.61–0.96) | 0.79 (0.56–0.92) | 0.82 (0.65–0.93) |
| NAIP + Clay | 0.87 (0.40–1.31) | 0.78 (0.43–1.05) | 0.77 (0.52–0.97) | 0.82 (0.63–0.96) | 0.87 (0.73–0.96) |
| NAIP + TESSERA | **1.01** (0.58–1.45) | 0.86 (0.60–1.11) | 0.79 (0.63–0.93) | 0.82 (0.68–0.92) | 0.82 (0.68–0.97) |
| *Detected (≥50% of pixels)* | | | | | |
| NAIP | **0.61** (0.35–0.80) | 0.73 (0.52–0.88) | **0.83** (0.70–0.92) | **0.94** (0.88–0.98) | **0.97** (0.93–1.00) |
| AEF | 0.46 (0.15–0.76) | 0.55 (0.24–0.79) | 0.66 (0.39–0.83) | 0.75 (0.52–0.91) | 0.77 (0.58–0.92) |
| Clay | 0.25 (0.10–0.41) | 0.40 (0.18–0.57) | 0.62 (0.39–0.79) | 0.79 (0.56–0.94) | 0.91 (0.72–1.00) |
| TESSERA | 0.44 (0.15–0.75) | 0.51 (0.22–0.76) | 0.60 (0.39–0.77) | 0.70 (0.53–0.84) | 0.75 (0.61–0.87) |
| NAIP + AEF | 0.55 (0.34–0.76) | **0.74** (0.53–0.88) | 0.77 (0.57–0.91) | 0.81 (0.58–0.96) | 0.88 (0.69–0.99) |
| NAIP + Clay | 0.48 (0.22–0.72) | 0.60 (0.31–0.81) | 0.71 (0.47–0.88) | 0.83 (0.62–0.96) | 0.93 (0.78–1.00) |
| NAIP + TESSERA | 0.50 (0.28–0.73) | 0.65 (0.45–0.82) | 0.74 (0.56–0.87) | 0.84 (0.68–0.94) | 0.83 (0.70–0.95) |

The NAIP baseline achieved the best overall performance, with an F1 of 0.74, recall of 0.78, and mIoU of 0.72. The three embedding-only models performed substantially worse. Clay was the strongest embedding-only model, and its precision (0.71) was comparable to that of the baseline, but its recall dropped to 0.65. AEF and TESSERA, by contrast, overestimated canopy extent, with area ratios of 1.25 and 1.35 and precision of only 0.57 and 0.50, respectively; TESSERA had the lowest F1 (0.57) and mIoU (0.58) of all models. The embeddings alone cannot delineate individual tree crowns as accurately as NAIP imagery, likely because their compact per-pixel descriptors lack the fine spatial detail needed to separate adjacent crowns.

Combining the embeddings with NAIP recovered most of the baseline performance. Notably, the overprediction observed for AEF and TESSERA alone was not carried over into the fusion models, suggesting that NAIP features dominate the spatial delineation of crowns while the embeddings mainly shift the balance between precision and recall. Specifically, all three fusion models lowered recall from 0.78 to 0.69–0.70, while NAIP + AEF and NAIP + Clay raised precision (0.76 and 0.74 vs. 0.70), whereas NAIP + TESSERA lowered it (0.67). As a result, none exceeded the NAIP baseline in F1 (0.68–0.72 vs. 0.74). Thus, none of the embeddings improved the aggregate segmentation accuracy, either alone or combined with NAIP, although their effects on recall, precision, and predicted canopy area differed.

The size-stratified results show that embedding effects depend on crown size (Table 4). All three embeddings performed worse than the NAIP baseline in every size class, but the gap narrowed as crowns became larger. The embedding-only models also failed on small crowns in different ways. AEF and TESSERA overestimated crown area for crowns smaller than 3 m and had low precision, whereas Clay underestimated it and detected only 25% of these crowns. In terms of F1, AEF was the strongest embedding-only model for crowns smaller than 5 m, while Clay was the strongest for crowns of 5 m and larger. Overall, segmentation accuracy increased with crown size for all models, but no fusion model exceeded the NAIP baseline in F1 by a meaningful margin in any size class. For crowns smaller than 3 m, NAIP alone achieved the highest recall (0.57) and detection rate (0.61), and fusion lowered recall to 0.44–0.54. For crowns of 8–15 m, fusion lowered recall from 0.81 to 0.70–0.73 and F1 from 0.85 to 0.79–0.80. Because the test crowns are concentrated in 21 of the 68 test patches, the 95% confidence intervals in Table 4 are wide, and several of these differences, particularly for the smallest and largest crowns, are not statistically distinguishable. The size-dependent results highlight the distinction between detecting canopy and accurately delineating individual crowns.

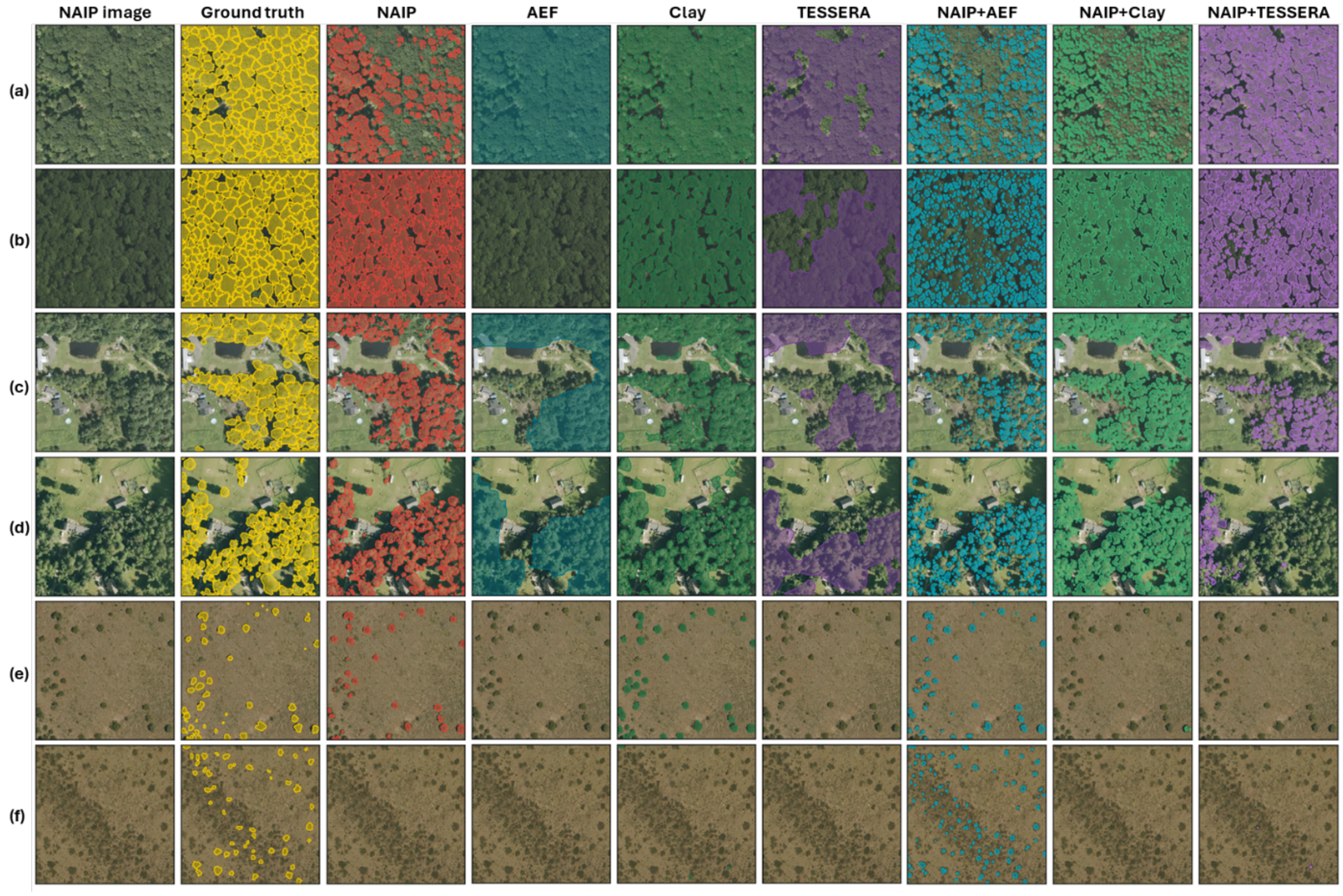

**Figure 7** Individual tree segmentation on six regions spanning closed to sparse canopy conditions. Columns from left to right show the NAIP true-color image, ground truth, and predictions from the four models. (a)–(f) indicate results of six representative regions.

The qualitative examples in Figure 7 further illustrate this distinction across different canopy conditions. In dense-canopy regions (Figure 7 (a) and (b)), where adjacent crowns have overlapping or weakly separated spectral and textural patterns, the predictions tend to reproduce the overall canopy extent more readily than the detailed boundaries of individual crowns. In sparse regions (Figure 7 (e) and (f)), the models perform differently on detecting small and isolated trees, leading to larger differences in recall and predicted area. The embedding-only models detected few or none of these trees, whereas NAIP-based approaches detected more of them. Across all crown-size classes, the three fusion models had lower recall than the NAIP baseline (e.g., 0.70–0.73 vs. 0.81 for 8–15 m crowns), and none exceeded it in F1 in any class. Within-crown precision was nearly unchanged (0.89 vs. 0.90 for 8–15 m crowns), so the higher overall precision of NAIP + AEF and NAIP + Clay (Table 3) came from fewer false positives away from reference crowns rather than from sharper crown boundaries . These examples are consistent with the quantitative results, showing that adding embeddings did not improve crown delineation. NAIP + AEF and NAIP + Clay traded recall for precision, whereas NAIP + TESSERA lowered both.

## 4.3 Crop Yield Prediction

In this case study, we evaluated the performance of two Earth embeddings relative to conventional remote sensing features for county-level soybean yield prediction. Annual soybean yield statistics for 12 states in the U.S. Midwest spanning 2017–2024 were obtained from the United States Department of Agriculture National Agricultural Statistics Service (USDA-NASS). For each year,

AEF and TESSERA embeddings were spatially aggregated to the county level by computing the mean value of each embedding dimension across soybean pixels within county boundaries, as identified by the Cropland Data Layer (CDL). This crop-specific masking procedure excluded non-soybean areas from the aggregation. Finally, the aggregated county-level embeddings were matched with the corresponding yield statistics to construct the dataset for soybean yield prediction.

For comparison, we followed a previous study (Ma et al., 2026) and organized hand-crafted remote sensing features using Landsat Collection 2 Tier 1 Land Surface Reflectance data from the same study period, including Landsat 7, 8, and 9. Six spectral bands were extracted, including Red, Blue, Green, NIR, Shortwave-Infrared-1 (SWIR1), and Shortwave-Infrared-2 (SWIR2), and two vegetation indices (VIs) were calculated as additional bands: Normalized Difference Vegetation Index (NDVI) and Green Chlorophyll Vegetation Index (GCVI). Next, a harmonic curve was fitted to the time series for each raw spectral and VI band in each county, and five phenology-based features were extracted from each fitted curve. In addition, weather data were downloaded from the ERA5-Land Climate Reanalysis dataset for each county, including monthly cumulative precipitation (PPT) and mean temperature (T) during the main growing season from May to September. Again, CDL was used to mask out non-soybean pixels during this process.

We used leave-one-year-out (LOYO) cross-validation to evaluate model performance. In each iteration, data from one year were held out as the test set, while data from all remaining years were used for training. This setup reflects a practical yield-prediction scenario in which training data are available only for certain years. Using Random Forest (RF) as the baseline model, we compared the performance when using AEF, TESSERA, and remote sensing and weather features as predictors for county-level soybean yield prediction (Table 5).

The AEF-based model generally had better performance than models using TESSERA embeddings or remote sensing features. When only using remote sensing features as predictors, the model achieved an overall $R^2$ of 0.703 and RMSE of 0.410 t/ha. Including additional weather features increased the overall $R^2$ slightly to 0.712 but still underperformed the AEF-based model that achieved an overall $R^2$ of 0.737 and RMSE of 0.386 t/ha. The TESSERA-based model produced a slightly higher $R^2$ of 0.725 than the remote-sensing-based model while still underperformed the AEF-based model.

This indicates that AEF embeddings capture richer and more comprehensive information related to crop growth and productivity. AEF embeddings integrate multiple complementary data modalities, including radar imagery, topographic information, gravity fields, and geocoded contextual features. Meanwhile, TESSERA embeddings are primarily learned from satellite imagery (Sentinel-1 and Sentinel-2), which capture a narrower set of information relevant to yield formation. These additional sources in AEF likely provide information that is not fully represented in optical remote sensing or weather data alone. For example, radar observations can be sensitive to canopy structure and surface moisture, topographic features can shape local drainage and erosion patterns, and geocoded context can capture spatially structured environmental or management-related variation. Together, these modalities may enable AEF embeddings to better characterize field-level growing conditions and productivity potential, thereby improving yield prediction performance.

**Table 5** Evaluation scores of county-level soybean yield prediction. The best result is in **bold,** and the second best is <u>underlined.</u>

| Year | N | AEF | | TESSERA | | RS | | RS + Weather | | RS + Weather + AEF | |
|---|---|---|---|---|---|---|---|---|---|---|---|
| | | R² | RMSE (t/ha) | R² | RMSE (t/ha) | R² | RMSE (t/ha) | R² | RMSE (t/ha) | R² | RMSE (t/ha) |
| 2017 | 802 | 0.684 ± 0.009 | 0.357 ± 0.005 | 0.688 ± 0.011 | 0.355 ± 0.006 | 0.683 ± 0.003 | 0.358 ± 0.002 | 0.720 ± 0.003 | 0.336 ± 0.002 | **0.764** ± 0.003 | **0.308** ± 0.002 |
| 2018 | 714 | 0.713 ± 0.004 | 0.363 ± 0.002 | 0.724 ± 0.002 | 0.356 ± 0.001 | 0.711 ± 0.002 | 0.364 ± 0.001 | 0.713 ± 0.010 | 0.363 ± 0.006 | **0.745** ± 0.003 | **0.342** ± 0.002 |
| 2019 | 638 | 0.671 ± 0.008 | 0.332 ± 0.004 | 0.695 ± 0.004 | 0.320 ± 0.002 | 0.618 ± 0.004 | 0.358 ± 0.002 | 0.672 ± 0.003 | 0.332 ± 0.001 | **0.724** ± 0.007 | **0.304** ± 0.004 |
| 2020 | 880 | 0.776 ± 0.002 | 0.316 ± 0.002 | 0.719 ± 0.002 | 0.354 ± 0.001 | 0.699 ± 0.002 | 0.366 ± 0.001 | 0.711 ± 0.002 | 0.359 ± 0.001 | **0.805** ± 0.002 | **0.295** ± 0.001 |
| 2021 | 752 | **0.799** ± 0.004 | **0.398** ± 0.004 | 0.701 ± 0.006 | 0.486 ± 0.005 | 0.747 ± 0.002 | 0.446 ± 0.002 | 0.683 ± 0.004 | 0.500 ± 0.003 | 0.767 ± 0.002 | 0.428 ± 0.002 |
| 2022 | 826 | 0.726 ± 0.004 | 0.450 ± 0.003 | 0.774 ± 0.003 | 0.409 ± 0.002 | 0.795 ± 0.001 | 0.390 ± 0.001 | 0.802 ± 0.003 | 0.383 ± 0.003 | **0.847** ± 0.003 | **0.337** ± 0.003 |
| 2023 | 756 | **0.743** ± 0.003 | **0.423** ± 0.002 | 0.733 ± 0.004 | 0.432 ± 0.003 | 0.654 ± 0.001 | 0.492 ± 0.001 | 0.687 ± 0.004 | 0.467 ± 0.003 | 0.736 ± 0.002 | 0.429 ± 0.002 |
| 2024 | 656 | 0.688 ± 0.002 | 0.432 ± 0.001 | 0.699 ± 0.005 | 0.424 ± 0.004 | 0.593 ± 0.005 | 0.493 ± 0.003 | 0.635 ± 0.005 | 0.467 ± 0.003 | **0.770** ± 0.002 | **0.370** ± 0.002 |
| **All** | **6,024** | 0.737 ± 0.001 | 0.386 ± 0.001 | 0.725 ± 0.002 | 0.395 ± 0.001 | 0.703 ± 0.001 | 0.410 ± 0.001 | 0.712 ± 0.002 | 0.404 ± 0.002 | **0.779** ± 0.001 | **0.354** ± 0.001 |

Notably, we found that conventional remote sensing features and AEF embeddings provide complementary information. By building an RF using AEF, remote sensing, and weather features as predictors, the model achieved the highest evaluation scores in six out of eight years and outperformed the second-best AEF-based RF by around 3% in terms of overall $R^2$. Again, AEF embeddings are designed as general-purpose geospatial representations that capture dynamics of Earth's surface. Therefore, they do not necessarily emphasize information most relevant to specific downstream tasks, such as yield prediction. Incorporating conventional hand-crafted features helps guide the model toward task-relevant signals by reweighting the information embedded across the multiple data modalities assimilated in AEF. This complementary information improves the model's ability to capture yield-related patterns, leading to enhanced prediction performance.

## 4.4 Forest Aboveground Biomass Density (AGBD) Mapping

In this case study, we investigated the spatiotemporal dynamics of aboveground biomass density (AGBD) in a tropical forest region of western Uganda characterized by documented forest degradation and reforestation. We compared models based on TESSERA and AEF embeddings (Figure 8 (a)–(f)) with a task-specific model trained on Landsat composite time-series model (Figure 8 (g)–(i)). We also evaluated Earth embeddings as complementary features by training models that combined the embeddings with Landsat composites (Figure 8 (j)–(o)).

We trained all models using quality-filtered footprint-level AGBD from GEDI L4A in the study area. The TESSERA-based model used 128-dimensional TESSERA embeddings as predictors in an XGBoost regressor, while the AEF-based model used 64-dimensional AEF embeddings in an XGBoost regressor. The Landsat-based model incorporated Landsat spectral bands, vegetation indices, and LandTrendr-derived change metrics from an eight-year retrospective window into an XGBoost regressor. The Landsat-based model followed Liang et al. (2023) and used systematic feature selection to reduce the initial 96 metrics to 24 predictors (Liang et al., 2026). No feature selection was implemented for TESSERA or AEF.

Our primary criterion for assessing model performance was the consistency of model estimates from independent forest inventory (FI) plots on the ground. To reduce sensitivity to geolocation mismatch between GEDI footprints and field plots, we applied all models at 30 m for 2017 and 2020 and aggregated predictions to 90 m before comparing them with FI-derived AGBD for each year and with the observed three-year biomass change. The model was trained using data from one year and evaluated in the other year. As shown in Figure 8, one clear advantage of the embedding-based models is their ability to support continuous, gap-free mapping, including in 2017, when the optical imagery only had limited clear-sky observations. This advantage likely reflects the inclusion of optical and Sentinel-1 SAR observations in both embedding products, which provide complementary information under cloudy conditions. On the other hand, the Landsat-based model used seasonal Landsat composites, especially including imagery from the limited hydric stress period following Liang et al. (2023). Such seasonal standardization can help reduce interannual variability associated with vegetation phenology and moisture conditions, which is important for consistent AGBD estimation in regions with distinct wet and dry seasons, including much of sub-Saharan Africa. However, this flexibility also makes the Landsat-based model more dependent on the availability of suitable clear-sky optical observations.

The generated maps also highlight the importance of post-processing and reveal differences in the spatial patterns of estimated AGBD across models. First, the top left inset maps in Figure 8 show that waterbodies were not automatically masked when embeddings were used as predictors, and biomass values were therefore estimated over non-forest areas. Applying a forest cover mask after model prediction is thus essential before estimating regional biomass totals. Secondly, across the forest landscape, all three models distinguished forest edges from more continuous forests, but the Landsat-based model captured stronger biomass contrasts across these gradients. The AEF-based model exhibited lower spatial contrast than the other two models, potentially reflecting its greater incorporation of neighborhood information through spatial attention in its encoder architecture. This may contribute to more spatially homogeneous biomass predictions, potentially reducing sensitivity to fine-scale forest structural heterogeneity. In contrast, TESSERA primarily relies on individual pixel spectral-temporal relationship; and Landsat-based models have neighborhood characteristics in certain features but primarily relied on pixel-level information as well. In the left and center columns in Figure 9, which compare model-derived AGBD estimates with FI-derived estimates for individual years, the Landsat-based model had a lower intercept and steeper fitted slope (Figure 9 (g)–(i)), suggesting less compression of the predicted biomass range, potentially indicating greater sensitivity to variation in high-biomass forests.

The three-year AGBD change estimates further highlight differences in the ability of the three approaches to capture temporal variations in forest biomass. The weaker spatial contrast in the AEF-based model, combined with the temporal transferability issues suggested by Ball et al. (2026), appears to limit its ability to reliably capture biomass change. This is especially important in the central part of the study area, where restoration projects have been active since 1995 and biomass gains are more gradual than abrupt loss signals. The embedding-based models show weaker agreement with FI-derived biomass changes than the Landsat-based model. This difference may partly reflect how temporal information is represented: while Earth embeddings summarize observations across broader temporal windows to support diverse downstream tasks, the Landsat-based model uses seasonally standardized Landsat composites tailored to periods with limited hydric stress, potentially improving sensitivity to gradual biomass changes by reducing phenological variability. As embedding-based models are less sensitive to subtle growth and recovery dynamics, their usefulness for change monitoring can be limited, even when they perform reasonably well for single-year biomass prediction.

More concerning is the difference in prediction bias between years (Figure 9). While the models exhibited relatively small biases in 2020, predictions for 2017 showed substantially larger biases. As 2017 predates GEDI observations, retrospectively applying models trained on later observations may introduce additional temporal inconsistencies in biomass estimates. These differences in year-specific bias can propagate directly into biomass change estimates, leading to substantial errors even when single-year biomass predictions show reasonable agreement with field observations. This highlights the importance of evaluating not only the accuracy of annual biomass estimates but also their temporal consistency when using Earth embeddings for forest carbon monitoring. We further tested using Earth embeddings as complementary features by training hybrid models that combined the embeddings with Landsat composites. Introducing Earth embeddings would mislead models to estimate biomass over waterbodies and lower the biomass contrasts captured by models. Although the hybrid models outperformed the Landsat-only models in 2017, this advantage did not hold in 2020, suggesting that Earth embeddings do not always complement remote sensing observations. Meanwhile, the hybrid models also failed to capture the biomass change in the three-year interval.

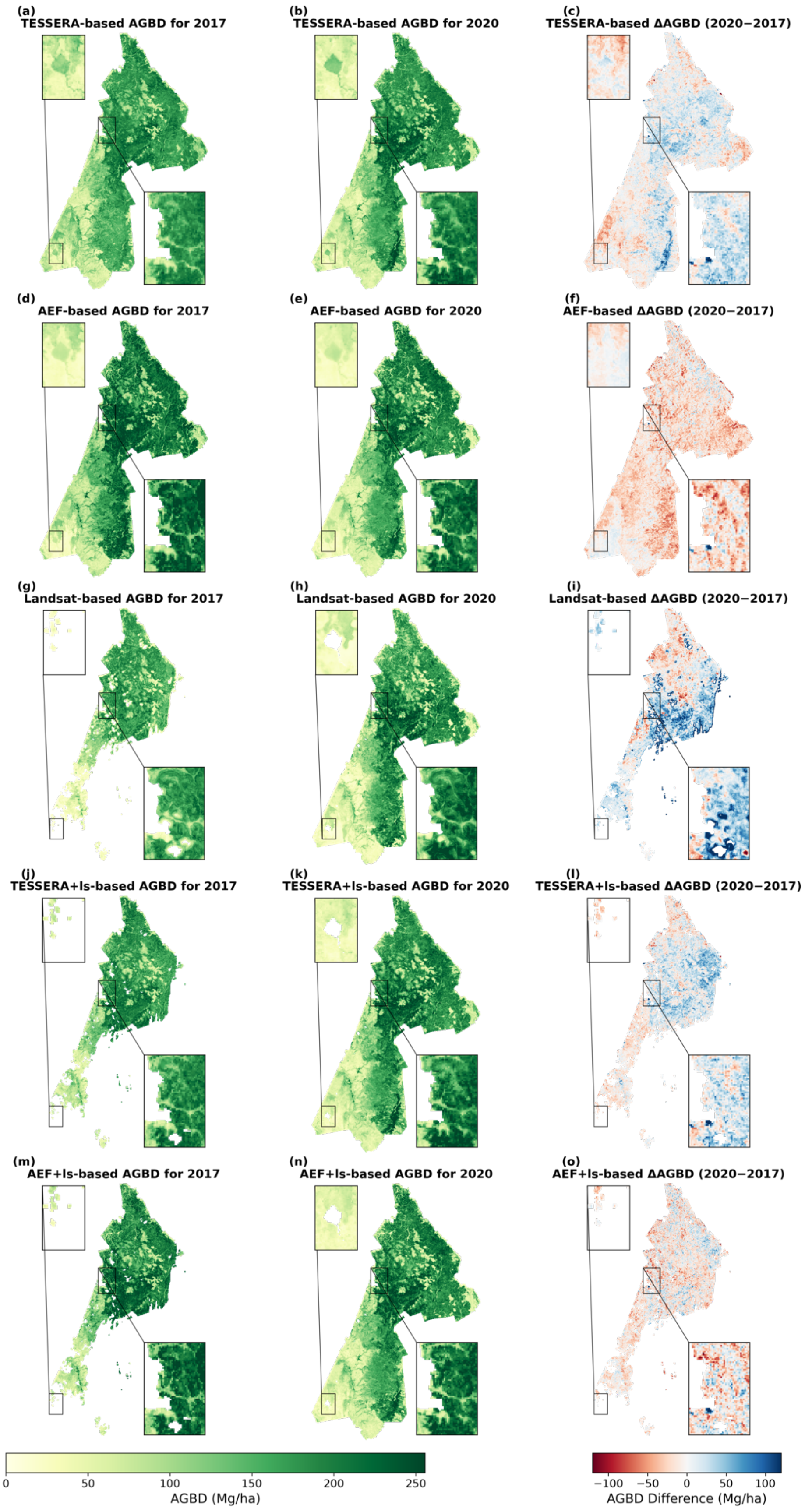
(a)
TESSERA-based AGBD for 2017
(b)
TESSERA-based AGBD for 2020
(c)
TESSERA-based ΔAGBD (2020−2017)
(d)
AEF-based AGBD for 2017
(e)
AEF-based AGBD for 2020
(f)
AEF-based ΔAGBD (2020−2017)
(g)
Landsat-based AGBD for 2017
(h)
Landsat-based AGBD for 2020
(i)
Landsat-based ΔAGBD (2020−2017)
(j)
TESSERA+ls-based AGBD for 2017
(k)
TESSERA+ls-based AGBD for 2020
(l)
TESSERA+ls-based ΔAGBD (2020−2017)
(m)
AEF+ls-based AGBD for 2017
(n)
AEF+ls-based AGBD for 2020
(o)
AEF+ls-based ΔAGBD (2020−2017)
0 50 100 150 200 250
AGBD (Mg/ha)
−100 −50 0 50 100
AGBD Difference (Mg/ha)

**Figure 8** Maps comparing AGBD estimates derived from models using (a)–(c) TESSERA, (d) – (f) AEF, (g)– (i) LS, (j)– (l) TESSERA + LS, and (m)– (o) AEF + LS only for 2017, 2020 and the AGBD change between these years across the Kibale National Park, Uganda. Top left inset maps show model estimates over a known waterbody in the study area. Bottom right inset maps highlight patterns over a known reforestation area. Each column corresponds to 2017, 2020, and the change between the two years, respectively.

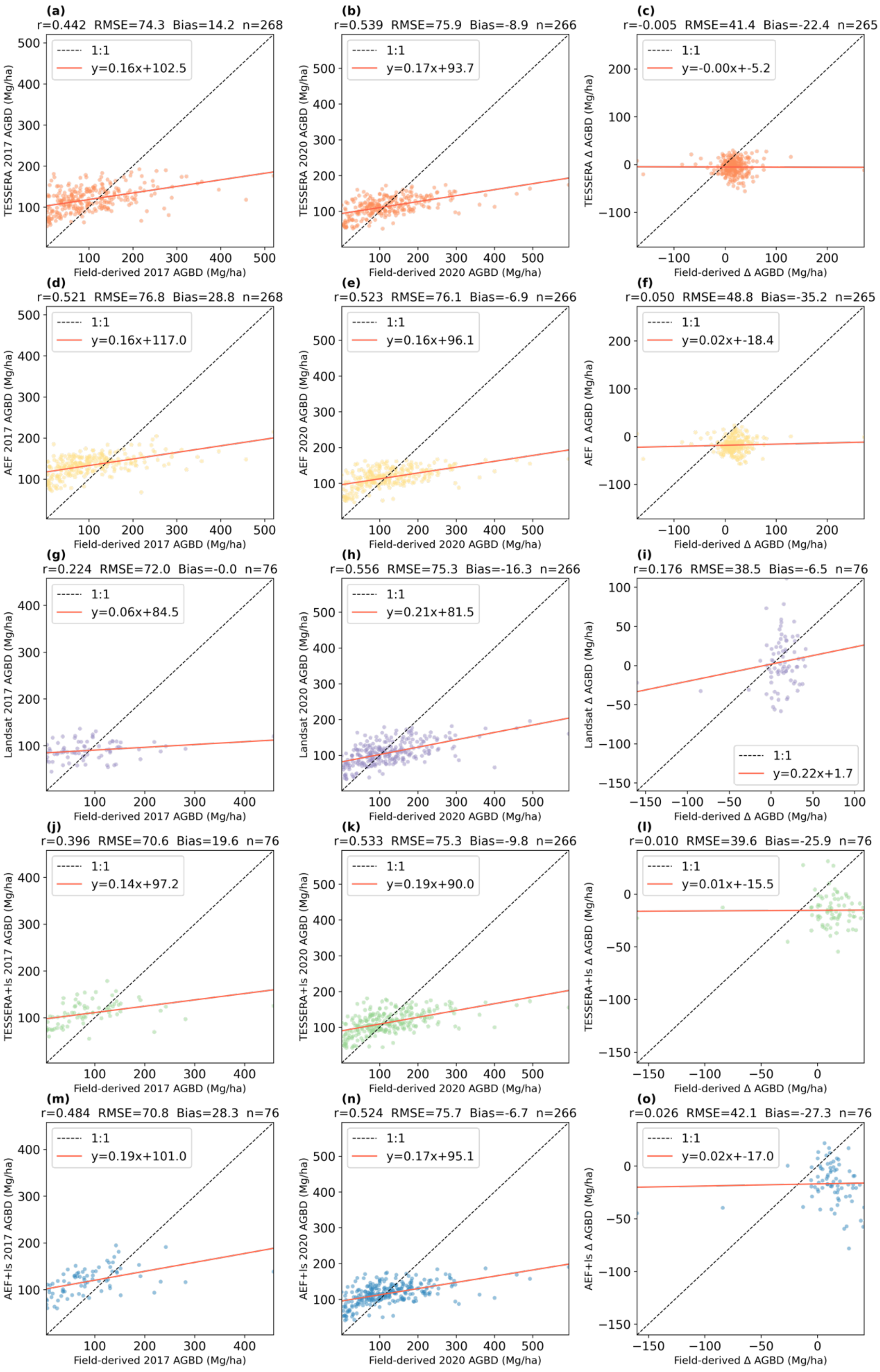

(a)
r=0.442 RMSE=74.3 Bias=14.2 n=268
1:1
y=0.16x+102.5
TESSERA 2017 AGBD (Mg/ha)
Field-derived 2017 AGBD (Mg/ha)
(b)
r=0.539 RMSE=75.9 Bias=-8.9 n=266
1:1
y=0.17x+93.7
TESSERA 2020 AGBD (Mg/ha)
Field-derived 2020 AGBD (Mg/ha)
(c)
r=-0.005 RMSE=41.4 Bias=-22.4 n=265
1:1
y=-0.00x+-5.2
TESSERA Δ AGBD (Mg/ha)
Field-derived Δ AGBD (Mg/ha)
(d)
r=0.521 RMSE=76.8 Bias=28.8 n=268
1:1
y=0.16x+117.0
AEF 2017 AGBD (Mg/ha)
Field-derived 2017 AGBD (Mg/ha)
(e)
r=0.523 RMSE=76.1 Bias=-6.9 n=266
1:1
y=0.16x+96.1
AEF 2020 AGBD (Mg/ha)
Field-derived 2020 AGBD (Mg/ha)
(f)
r=0.050 RMSE=48.8 Bias=-35.2 n=265
1:1
y=0.02x+-18.4
AEF Δ AGBD (Mg/ha)
Field-derived Δ AGBD (Mg/ha)
(g)
r=0.224 RMSE=72.0 Bias=-0.0 n=76
1:1
y=0.06x+84.5
Landsat 2017 AGBD (Mg/ha)
Field-derived 2017 AGBD (Mg/ha)
(h)
r=0.556 RMSE=75.3 Bias=-16.3 n=266
1:1
y=0.21x+81.5
Landsat 2020 AGBD (Mg/ha)
Field-derived 2020 AGBD (Mg/ha)
(i)
r=0.176 RMSE=38.5 Bias=-6.5 n=76
1:1
y=0.22x+1.7
Landsat Δ AGBD (Mg/ha)
Field-derived Δ AGBD (Mg/ha)
(j)
r=0.396 RMSE=70.6 Bias=19.6 n=76
1:1
y=0.14x+97.2
TESSERA+ls 2017 AGBD (Mg/ha)
Field-derived 2017 AGBD (Mg/ha)
(k)
r=0.533 RMSE=75.3 Bias=-9.8 n=266
1:1
y=0.19x+90.0
TESSERA+ls 2020 AGBD (Mg/ha)
Field-derived 2020 AGBD (Mg/ha)
(l)
r=0.010 RMSE=39.6 Bias=-25.9 n=76
1:1
y=0.01x+-15.5
TESSERA+ls Δ AGBD (Mg/ha)
Field-derived Δ AGBD (Mg/ha)
(m)
r=0.484 RMSE=70.8 Bias=28.3 n=76
1:1
y=0.19x+101.0
AEF+ls 2017 AGBD (Mg/ha)
Field-derived 2017 AGBD (Mg/ha)
(n)
r=0.524 RMSE=75.7 Bias=-6.7 n=266
1:1
y=0.17x+95.1
AEF+ls 2020 AGBD (Mg/ha)
Field-derived 2020 AGBD (Mg/ha)
(o)
r=0.026 RMSE=42.1 Bias=-27.3 n=76
1:1
y=0.02x+-17.0
AEF+ls Δ AGBD (Mg/ha)
Field-derived Δ AGBD (Mg/ha)

**Figure 9** Model performance evaluated against field-derived AGBD estimates for 2017 and 2020, and the change in AGBD between these years in Kibale National Park, Uganda. Panels show AGBD predictions derived from estimates by models using (a)–(c) TESSERA, (d) – (f) AEF, (g)–(i) LS, (j)– (l) TESSERA + LS, and (m)– (o) AEF + LS. Each column, from left to right, corresponds to 2017, 2020, and the change between the two years (2020 minus 2017), respectively.

# 5 Discussion

## 5.1 Advantages of Earth embeddings

In environmental remote sensing, building task-specific features often requires extensive preprocessing and feature engineering, including sensor and modality selection, uncertainty and noise control, spatial or temporal aggregation, data harmonization, and regularization. Although geospatial computing platforms such as Google Earth Engine and Microsoft's Planetary Computer provide reliable infrastructure for these workflows, they still require substantial domain knowledge and coding effort.

Earth embeddings, as task-agnostic descriptors of the Earth's surface, can substantially reduce this burden. Practitioners can use embeddings directly for downstream tasks without separately processing each modality or developing custom data harmonization pipelines. For example, in the soybean yield prediction case study, we independently downloaded Landsat time series and ERA5 weather data. By contrast, Earth embeddings were downloaded directly for the study area and used with only minor preprocessing (cropland masking and spatial aggregation). Moreover, precomputed embeddings are easier to download and access. Our study showed that extracting the county-level Landsat time series across the 12 U.S. Midwestern states required approximately 436,000 Earth Engine Compute Unit (EECU) seconds, whereas retrieving the corresponding AEF embeddings over the same region required only about 69,000 EECU seconds—a more than six-fold reduction.

Another advantage of Earth embeddings is their ability to support continuous, cloud-free mapping. By fusing multi-modal observations, cloud-prone sources such as optical spectral imagery can be complemented by cloud-robust sources (SAR), producing seamless embeddings across the Earth's surface. As demonstrated in AGBD mapping, both AEF and TESSERA embeddings enabled continuous biomass mapping in both cloud-heavy (2017) and cloud-light years (2020).

In addition, as location-based representations, Earth embeddings provide complementary spatial and environmental context that is not fully captured by spectral observations alone. In HSI classification, Earth embeddings improved robustness under spatial separation between training and testing samples. When training and testing samples came from geographically disjoint areas within the scene, TESSERA embeddings improved both accuracy and stability, acting as a spatially consistent prior that compensated for the limited transferability of locally calibrated spectral features.

Finally, Earth embeddings can serve as a form of prediction regularization. Because they encode broader spatial and temporal context, they can help suppress overprediction. For example, in individual tree crown segmentation, Earth embeddings helped temper canopy overprediction while supporting the detection of small, weakly evidenced targets that occupied only a few pixels in the primary imagery.

## 5.2 Limits of task-agnostic embeddings

Although producers of Earth embeddings often claim superior performance, our case studies did not show consistent improvements from using these embeddings. Instead, the strongest overall performance under standard evaluation settings mostly came from baselines using only the primary imagery, while using embeddings degraded accuracy as often as it improved it. Based on our case studies, we summarized the limits of task-agnostic embeddings as below:

**Information redundancy**. Earth embeddings can contain redundant information relative the demand of downstream tasks. In HSI classification, fine-grained urban classes are primarily distinguished by detailed spectral signatures already captured by the hyperspectral data. The broader land-cover information compressed into Earth embeddings may therefore duplicate the primary input, while the additional feature dimensions can introduce noise when labels are limited. In individual tree crown segmentation, 10-m embeddings are coarser than many individual tree crowns and much coarser than the boundaries between adjacent crowns. Consequently, they cannot provide the fine-scale evidence required for crown delineation. Meanwhile, embeddings can take in irrelevant information. In AGBD mapping, embedding-based models consistently map biomass over waterbodies, which RS-based models can easily avoid. For AEF specifically, the incorporation of GEDI information into the embeddings may introduce redundancy when GEDI-derived metrics, such as AGBD, are used as training targets.

**Varying temporal stability.** The temporal stability of Earth embeddings can vary. Although the AEF-based model achieved the best overall performance in soybean yield prediction, it did not consistently outperform TESSERA-based or RS-based models in all eight testing years (Table 5). This suggests that AEF embeddings do not always capture the most predictive information related to the task each year. Since input data availability and quality may vary across years, annual embeddings may underweight key seasonal signals in certain testing years, such as peak-season crop greenness or late-season heat stress. Similarly, in AGBD mapping, embedding-based models performed comparably to or better than RS-based models in individual test years but were less sensitive to spatial and temporal variability in biomass and failed to capture biomass change over the three-year interval.

**Spatial mismatch.** The fixed spatial resolutions of Earth embeddings can affect their applicability. For individual tree crown segmentation, the model must identify individual objects and accurately delineate their boundaries. This creates a stronger scale mismatch between Earth embeddings and the downstream target. The broader context encoded by embeddings can provide useful cues about vegetation occurrence, particularly for small or isolated crowns, but it cannot substitute for the fine-scale spatial information needed to separate adjacent crowns. Similarly, in AGBD mapping, maps generated using AEF-based and TESSERA-based models show lower spatial contrast, suggesting that embedding-based models may smooth fine-scale heterogeneity.

**Task-dependent performance.** The effectiveness of Earth embeddings depends strongly on the type of information required by the downstream task. For example, in hyperspectral image classification, both spectral signatures and spatial context are fundamental for distinguishing land-cover classes, particularly when neighboring materials exhibit similar spectral responses or when class boundaries are spatially structured. Earth embeddings can complement these task-specific features by encoding a wider environmental, temporal, and spatial context that may not be fully captured within a local HSI patch. Their benefit, however, depends on whether this contextual information is sufficiently complementary to the fine-grained spectral–spatial representation learned from the HSI.

**Potential circularity.** Although AEF embeddings demonstrated promising performance in soybean yield prediction, potential circularity in model evaluation warrants careful consideration. Specifically, geolocated Wikipedia text used during AEF pre-training may contain agricultural production information from the evaluation years, potentially introducing information leakage. Similarly, GEDI canopy height data were incorporated into AEF pre-training and subsequently used as ground truth in the AGBD mapping case study. Such overlap between pre-training data and evaluation targets raises concerns about potential circularity, which could lead to inflated estimates of embedding performance on public benchmark datasets. Supporting this concern, a related study found that AEF-based models did not outperform Landsat-based models in yield prediction when evaluated using private field-level yield data (Ma et al., 2026).

## 5.3 Provenance and Openness

Another point that practitioners need to consider is the provenance and openness of different Earth embedding models and products (Table 6). The widely used AEF embeddings are publicly available via Google Earth Engine, while its model structure and model weights are not publicly available. OlmoEarth and Clay, on the other hand, released the source code and model weights, but the inner embeddings are not ready-to-use but need to be generated on demand.

The practical consequences of these provenance choices are visible in the case studies for hyperspectral image classification and individual tree segmentation. The closed AEF product could only be consumed as released, at fixed 64-dimensional, 10 m, annual granularity queried from Google Earth Engine. In contrast, the open Clay encoder was re-run on custom inputs, a 2018 Sentinel-2 composite for the hyperspectral experiment and native NAIP imagery for crown segmentation, where it yielded an effective token footprint of approximately 4.8 m, and the open TESSERA pipeline allowed embeddings to be generated for regions and years of interest, ensuring temporal alignment with the hyperspectral acquisition. Thus, openness determines not only reproducibility but also whether an embedding can be adapted in space, time, and sensor to the requirements of a downstream task.

**Table 6** Provenance and openness of the representative Earth embedding models and products

| Product | Model | Architecture | Training objective | Weights | Code | Reprod. |
|---|---|---|---|---|---|---|
| Google Satellite Embedding | AlphaEarth Foundations | Space-Time-Precision (STP) encoders | input reconstruction + thematic product reconstruction + text contrastive modeling | ✗ | ✗ | ✗ |
| TESSERA | TESSERA | dual-branch transformer (S1 + S2 time series) → MLP fusion | contrastive, on augmented pixel time series | ✓ | ✓ | ✓ |
| OlmoEarth | OlmoEarth v1–v1.2 (Nano–Large) | ViT over multimodal image time series; tokens carry location, time, sensor | latent masked image modeling with EO-specific masking and loss | ✓ | ✓ | ✓ |
| ESD | ESDNet | spatiotemporal fusion → latent encoding + FSQ → multitask heads | SDC30 reconstruction + multitask supervision | ✓ | ✓ | partial |
| Major TOM embeddings | SSL4EO-ResNet50 (DINO / MoCo), DINOv2, SigLIP-SO400M | ResNet-50; ViT | inherited from each encoder | ✓ | ✓ | ✓ |
| Clay | Clay v1.5 | ViT-L MAE + dynamic embedding block, GSD-scaled position encoding | 95% masked reconstruction + 5% DINOv2 distillation | ✓ | ✓ | ✓ |
| Copernicus-Embed | Copernicus-FM | ViT + dynamic hypernetworks, Fourier metadata encoding | masked image modeling + continual distillation | ✓ | ✓ | ✓ |

## 5.4 Practical implications

The advantages, limits, and openness of Earth embeddings indicates how embedding products connect to the broader trajectory of GFMs. Embeddings are essentially a deployment interface for such models. GFMs are trained once at scale, and the learned geospatial representations are distributed as fixed, analysis-ready vectors. Some trade the representational control of fine-tuning for accessibility, while freezing design choices such as spatial support and annual compositing that our case studies identify as the main limiting factors. The others, while providing open access to their architectures and weights, may still encode redundant or irrelevant information that can negatively affect downstream task performance.

We recommend remote sensing practitioners carefully evaluate Earth embeddings' suitability for specific tasks rather than assuming superior performance by default. Embeddings should be treated as candidate features whose value depends on the relationship between what they encode and what the task requires. When the primary sensor already captures the key discriminative information, task-agnostic embeddings may offer little additional benefit and may instead introduce redundancy. The temporal stability of Earth embeddings should also be carefully evaluated. For multi-year tasks, a single Earth embedding product may fail to deliver the best performance consistently across all testing years.

In addition, future work on Earth embeddings for environmental monitoring should closely examine how spatial and temporal information is encoded and whether users can better align embeddings with ecologically meaningful time windows. Ongoing developments, including sensor-agnostic encoders, multi-scale representations, and sub-annual releases, target precisely these constraints, so the range of tasks served by fixed embeddings is likely to expand while boundary-sensitive tasks such as crown delineation will continue to require the primary observations or the model itself. GFMs and embedding products are therefore best viewed as co-evolving deployment routes, to be chosen according to whether the embedding's scale and semantics match the downstream target.

Moreover, with the emergence of agentic AI (Lu et al., 2024), the benefits of Earth embeddings being easy to work with would presumably reduce, since AI agents and AI scientists would be able to automatically design, extract, and harmonize features from multi-modal EO that are suited to a task. Therefore, Earth embeddings should not be viewed as a simple replacement for conventional EO data. Instead, they should be used strategically, for example, as representations of the Earth surface that provide valuable prior information for downstream geospatial tasks with the potential to complete existing remote sensing features.

# 6 Conclusion

The development and availability of Earth embedding products have expanded rapidly since 2024. Earth embeddings show considerable potential to address key limitations of traditional EO, including observation gaps and the substantial effort required for data harmonization and preprocessing. In this review, we introduce the theoretical foundations of Earth embeddings and provide an overview of several representative products. We further review representative applications of Earth embeddings in environmental remote sensing. To date, most studies have focused on land-cover classification, while the potential of Earth embeddings remains largely unexplored across many other environmental applications. In addition, we conduct case studies and systematically assess the performance of Earth embeddings in four downstream tasks, including hyperspectral imagery classification, individual tree crown segmentation, soybean yield prediction, and above-ground biomass density mapping.

Building on the literature review and empirical comparisons, we identify several advantages of Earth embeddings. However, current embedding products do not consistently outperform conventional hand-crafted remote sensing features across all tasks, regions, or evaluation settings. Our case studies suggest that embeddings and hand-crafted features capture partly distinct and complementary information, which leads to differences in performance across years, regions, and task types. The inter-comparison of embedding products further highlights their varying strengths. No single Earth embedding product performs best across all tasks.

Overall, Earth embeddings should not be viewed as a direct replacement for conventional EO data, but rather as representations that provide stable information across space and time to support downstream geospatial tasks. Practitioner can benefit from retaining a EO baseline evaluated under the same protocol. Such comparisons provide the most direct evidence for assessing the added value of Earth embeddings.

**Data Availability**

The AlphaEarth Foundations (AEF) annual embeddings are available in the Google Earth Engine Data Catalog as the "GOOGLE/SATELLITE_EMBEDDING/V1/ANNUAL" image collection. TESSERA embeddings are available through the geotessera Python library (https://github.com/ucam-eo/geotessera). Clay embeddings were generated with the openly released Clay v1.5 model weights (https://huggingface.co/made-with-clay/Clay; Clay Foundation, 2024). The Landsat imagery and the USGS DEM datasets are available on Google Earth Engine (https://earthengine.google.com/). The Augsburg hyperspectral image and land-cover reference data are publicly available at https://github.com/danfenghong/ISPRS_S2FL. The Sentinel-2 Level-2A imagery used to generate the Clay embeddings is freely available from the Copernicus Data Space Ecosystem (https://dataspace.copernicus.eu/). NAIP imagery is publicly available from the U.S. Geological Survey (https://doi.org/10.5066/F7QN651G). The USDA county-level yield statistics are available at USDA NASS Quick Stats (https://quickstats.nass.usda.gov/). GEDI L4A footprint-level aboveground biomass density data (Version 2.1) are available from the ORNL DAAC (https://doi.org/10.3334/ORNLDAAC/2056).

**Acknowledge**

Funding was provided by NASA Acres (NASA Applied Sciences Grant No. 80NSSC23M0034, sub-award 124245-Z6512205 to D.B.L.). Support for this research was also provided by the University of Wisconsin-Madison, Office of the Vice Chancellor for Research and Graduate Education with funding from the Wisconsin Alumni Research Foundation, and Earth and Environmental Systems Modeling Program, Office of Biological and Environmental Research of the U.S. Department of Energy Office of Science (DE-SC0025251). Any opinions, findings, and conclusions or recommendations expressed in this study are those of the authors and do not necessarily reflect the views of NASA.

## References

Albughdadi, M., Antonacci, M., Baousis, V., Fornari, F., Kaprol, T., Pisa, C., 2025. Unsupervised Change Detection Using Sentinel-1 and Sentinel-2 Imagery with the Clay Foundation Model: A Case Study of Flood-Affected Areas in Valencia Spain. Presented at the EGU General Assembly 2025, Vienna, Austria, EGU25-8272. https://doi.org/10.5194/egusphere-egu25-8272

Ball, J.G.C., Wicklein, J.A., Feng, Z., Knezevic, J., Jaffer, S., Madhavapeddy, A., Atzberger, C., Dalponte, M., Coomes, D.A., 2026. Geospatial foundation models enable data-efficient tree species mapping in temperate mountain forests. Sci. Remote Sens. 14, 100466. https://doi.org/10.1016/j.srs.2026.100466

Bastani, F., Wolters, P., Gupta, R., Ferdinando, J., Kembhavi, A., 2023. SatlasPretrain: A Large-Scale Dataset for Remote Sensing Image Understanding, in: 2023 IEEE/CVF International Conference on Computer Vision (ICCV). Presented at the 2023 IEEE/CVF International Conference on Computer Vision (ICCV), pp. 16726–16736. https://doi.org/10.1109/ICCV51070.2023.01538

Benavides-Martinez, I.F., Guthrie, J., Arias, J.E., Garces-Gomez, Y.A., Guzman-Alvis, A.I., Portilla-Cabrera, C.V., Mondal, S., Allyn, A.J., Ganguly, A.R., 2026. What on Earth is AlphaEarth? Hierarchical structure and functional interpretability for global land cover. https://doi.org/10.48550/arXiv.2603.16911

Blasch, G., Alemayehu, Y., Lesne, L., Wolter, J., Taymans, M., Tesfaye, T., Negash, T., Andulalem, M., Gutu, K., Debela, M., Eshetu, Z., Tesfaye, K., Mottaleb, K., Defourny, P., Hodson, D.P., 2024. Ethiopian Crop Type 2020 (EthCT2020) dataset: Crop type data for environmental and agricultural remote sensing applications in complex Ethiopian smallholder wheat-based farming systems (Meher season 2020/21). Data Brief 54, 110427. https://doi.org/10.1016/j.dib.2024.110427

Brown, C.F., Kazmierski, M.R., Pasquarella, V.J., Rucklidge, W.J., Samsikova, M., Zhang, C., Shelhamer, E., Lahera, E., Wiles, O., Ilyushchenko, S., Gorelick, N., Zhang, L.L., Alj, S., Schechter, E., Askay, S., Guinan, O., Moore, R., Boukouvalas, A., Kohli, P., 2025. AlphaEarth Foundations: An embedding field model for accurate and efficient global mapping from sparse label data. https://doi.org/10.48550/arXiv.2507.22291

Buchanan, S., Pai, D., Wang, P., Ma, Y., 2026. Principles and Practice of Deep Representation Learning: or a Mathematical Theory of Memory. https://doi.org/10.48550/arXiv.2606.06624

Castaldi, F., Syu, C.-H., Valdez, M.C., Yen, C.-C., Chen, C.-F., Bertinaria, F., Toscano, P., 2026. Monitoring soil organic carbon from earth observation in the era of cover cropping and no-tillage. Smart Agric. Technol. 13, 101935. https://doi.org/10.1016/j.atech.2026.101935

Chen, S., Wang, J., Yuan, S., Li, J., Xia, Y., Liao, Y., Wei, J., Yuan, J., Xu, X., Zhu, X., Zhu, P., Zhang, H., Zhou, Y., Fu, H., Huang, H., Chen, B., Dai, F., Gong, P., 2026. Democratizing planetary-scale analysis: an ultra-lightweight Earth embedding database for accurate and flexible global land monitoring. Earth Syst. Sci. Data 18, 5375–5398. https://doi.org/10.5194/essd-18-5375-2026

Cheng, Y., Zhu, Q., Fan, L., 2026. From Landslide Conditioning Factors to Satellite Embeddings: Evaluating the Utilisation of Google AlphaEarth for Landslide Susceptibility Mapping using Deep Learning. https://doi.org/10.48550/arXiv.2601.07268

Clay Foundation, 2024. Clay Foundation Model v1.5. https://clay-foundation.github.io/model/

Cong, Y., Khanna, S., Meng, C., Liu, P., Rozi, E., He, Y., Burke, M., Lobell, D.B., Ermon, S., 2022. SatMAE: Pre-training Transformers for Temporal and Multi-Spectral Satellite Imagery. Adv. Neural Inf. Process. Syst. 35, 197–211. https://doi.org/10.52202/068431-0015

Czerkawski, M., Kluczek, M., Bojanowski, J.S., 2024. Global and Dense Embeddings of Earth: Major TOM Floating in the Latent Space. https://doi.org/10.48550/arXiv.2412.05600

Deng, J., Dong, W., Socher, R., Li, L.-J., Li, K., Fei-Fei, L., 2009. ImageNet: A large-scale hierarchical image database, in: 2009 IEEE Conference on Computer Vision and Pattern Recognition. Presented at the 2009 IEEE Conference on Computer Vision and Pattern Recognition (CVPR), pp. 248–255. https://doi.org/10.1109/CVPR.2009.5206848

Ding, L., Hong, D., Zhao, M., Chen, H., Li, C., Deng, J., Yokoya, N., Bruzzone, L., Chanussot, J., 2025. A Survey of Sample-Efficient Deep Learning for Change Detection in Remote Sensing: Tasks, strategies, and challenges. IEEE Geosci. Remote Sens. Mag. 13, 164–189. https://doi.org/10.1109/MGRS.2025.3533605

Fang, H., Stewart, A.J., Corley, I., Zhu, X.X., Azizpour, H., 2026. Earth Embeddings as Products: Taxonomy, Ecosystem, and Standardized Access. https://doi.org/10.48550/arXiv.2601.13134

Feng, W., Guan, F., Sun, C., Xu, W., 2024. Feature-Differencing-Based Self-Supervised Pre-Training for Land-Use/Land-Cover Change Detection in High-Resolution Remote Sensing Images. Land 13, 927. https://doi.org/10.3390/land13070927

Feng, Z., Atzberger, C., Jaffer, S., Knezevic, J., Sormunen, S., Young, R., Lisaius, M.C., Immitzer, M., Jackson, T., Ball, J., Coomes, D.A., Madhavapeddy, A., Blake, A., Keshav, S., 2026. TESSERA: Temporal Embeddings of Surface Spectra for Earth Representation and Analysis, in: Proceedings of the IEEE/CVF Conference on Computer Vision and Pattern Recognition (CVPR). Presented at the IEEE/CVF Conference on Computer Vision and Pattern Recognition (CVPR), pp. 34818–34831.

Fisette, T., Rollin, P., Aly, Z., Campbell, L., Daneshfar, B., Filyer, P., Smith, A., Davidson, A., Shang, J., Jarvis, I., 2013. AAFC annual crop inventory, in: 2013 Second International Conference on Agro-Geoinformatics (Agro-Geoinformatics). Presented at the 2013 Second International Conference on Agro-Geoinformatics (Agro-Geoinformatics), pp. 270–274. https://doi.org/10.1109/Argo-Geoinformatics.2013.6621920

Fuller, A., Millard, K., Green, J.R., 2022. SatViT: Pretraining Transformers for Earth Observation. IEEE Geosci. Remote Sens. Lett. 19, 1–5. https://doi.org/10.1109/LGRS.2022.3201489

Fuller, A., Millard, K., Green, J.R., 2023. CROMA: Remote Sensing Representations with Contrastive Radar-Optical Masked Autoencoders. Adv. Neural Inf. Process. Syst. 36, 5506–5538. https://doi.org/10.52202/075280-0241

Gilch, L., Wittmann, I., Nitsche, M., Jakubik, J., Ewald, A., Brunschwiler, T., 2026. How to Embed Matters: Evaluation of EO Embedding Design Choices. https://doi.org/10.48550/arXiv.2603.10658

Gong, W., Srivastava, U., Wang, Y., Jia, Y., Wu, Q., Bai, W., Yang, Y., Huang, X., Ye, X., 2026. Earth Embeddings Reveal Diverse Urban Signals from Space. https://doi.org/10.48550/arXiv.2604.03456

Guan, K., Jin, Z., Peng, B., Tang, J., DeLucia, E.H., West, P.C., Jiang, C., Wang, S., Kim, T., Zhou, W., Griffis, T., Liu, L., Yang, W.H., Qin, Z., Yang, Q., Margenot, A., Stuchiner, E.R., Kumar, V., Bernacchi, C., Coppess, J., Novick, K.A., Gerber, J., Jahn, M., Khanna, M., Lee, D., Chen, Z., Yang, S.-J., 2023. A scalable framework for quantifying field-level agricultural carbon outcomes. Earth-Sci. Rev. 243, 104462. https://doi.org/10.1016/j.earscirev.2023.104462

He, K., Chen, X., Xie, S., Li, Y., Dollár, P., Girshick, R., 2022. Masked Autoencoders Are Scalable Vision Learners, in: 2022 IEEE/CVF Conference on Computer Vision and Pattern Recognition (CVPR). Presented at the 2022 IEEE/CVF Conference on Computer Vision and Pattern Recognition (CVPR), pp. 15979–15988. https://doi.org/10.1109/CVPR52688.2022.01553

Heiman, A., 2026. Characterizing Brazilian Atlantic Forest Restoration Outcomes with Geospatial AlphaEarth Embeddings. https://doi.org/10.48550/arXiv.2605.05547

Herzog, H., Bastani, F., Zhang, Y., Tseng, G., Redmon, J., Sablon, H., Park, R., Morrison, J., Buraczynski, A., Farley, K., Hansen, J., Howe, A., Johnson, P.A., Otterlee, M., Schmitt, T., Pitelka, H., Daspit, S., Ratner, R., Wilhelm, C., Wood, S., Jacobi, M., Kerner, H., Shelhamer, E., Farhadi, A., Krishna, R., Beukema, P., 2026. OlmoEarth: Stable Latent Image Modeling for Multimodal Earth Observation, in: Proceedings of the IEEE/CVF Conference on Computer Vision and Pattern Recognition (CVPR). Presented at the IEEE/CVF Conference on Computer Vision and Pattern Recognition (CVPR), pp. 34806–34817.

Hong, D., Zhang, B., Li, X., Li, Y., Li, C., Yao, J., Yokoya, N., Li, H., Ghamisi, P., Jia, X., Plaza, A., Gamba, P., Benediktsson, J.A., Chanussot, J., 2024. SpectralGPT: Spectral Remote Sensing Foundation Model. IEEE Trans. Pattern Anal. Mach. Intell. 46, 5227–5244. https://doi.org/10.1109/TPAMI.2024.3362475

Hou, S., Jiao, H., Qing, Y., Xu, Q., Liu, Z., Xie, L., Xu, Z., Wang, Z., Tang, S., Guan, X., Wu, H., 2026. Can AlphaEarth foundations redefine the paradigm of gridded population mapping? A systematic evaluation across 18 global cities and large-scale mapping applications. Remote Sens. Environ. 347, 115690.

Hu, J., Liu, R., Hong, D., Camero, A., Yao, J., Schneider, M., Kurz, F., Segl, K., Zhu, X.X., 2023. MDAS: a new multimodal benchmark dataset for remote sensing. Earth Syst. Sci. Data 15, 113–131. https://doi.org/10.5194/essd-15-113-2023

Jakubik, J., Roy, S., Phillips, C.E., Fraccaro, P., Godwin, D., Zadrozny, B., Szwarcman, D., Gomes, C., Nyirjesy, G., Edwards, B., Kimura, D., Simumba, N., Chu, L., Mukkavilli, S.K., Lambhate, D., Das, K., Bangalore, R., Oliveira, D., Muszynski, M., Ankur, K., Ramasubramanian, M., Gurung, I., Khallaghi, S., Li, H., Cecil, M., Ahmadi, M., Kordi, F., Alemohammad, H., Maskey, M., Ganti, R., Weldemariam, K., Ramachandran, R., 2023. Foundation Models for Generalist Geospatial Artificial Intelligence. https://doi.org/10.48550/arXiv.2310.18660

Jakubik, J., Yang, F., Blumenstiel, B., Scheurer, E., Sedona, R., Maurogiovanni, S., Bosmans, J., Dionelis, N., Marsocci, V., Kopp, N., Ramachandran, R., Fraccaro, P., Brunschwiler, T., Cavallaro, G., Bernabe-Moreno, J., Longépé, N., 2025. TerraMind: Large-Scale Generative Multimodality for Earth Observation, in: 2025 IEEE/CVF International Conference on Computer Vision (ICCV). Presented at the 2025 IEEE/CVF International Conference on Computer Vision (ICCV), pp. 7383–7394. https://doi.org/10.1109/ICCV51701.2025.00693

Jean, N., Wang, S., Samar, A., Azzari, G., Lobell, D., Ermon, S., 2019. Tile2Vec: Unsupervised Representation Learning for Spatially Distributed Data. Proc. AAAI Conf. Artif. Intell. 33, 3967–3974. https://doi.org/10.1609/aaai.v33i01.33013967

Jin, C., Jiang, X., Wen, L., Wu, C., Xu, X., Jiao, J., 2026. Assessing the Utility of Satellite Embedding Features for Biomass Prediction in Subtropical Forests with Machine Learning. Remote Sens. 18, 436. https://doi.org/10.3390/rs18030436

Kawakami, Y., Cayan, D., Liu, D., Ma, K.-L., Corringham, T., 2026. DELUGE: Towards Continental-Scale Daily Pluvial Flood Damage Prediction via Interpretable Conditioning on Foundation Model Embeddings. https://doi.org/10.48550/arXiv.2607.16050

Kerner, H., Nakalembe, C., Yang, A., Zvonkov, I., McWeeny, R., Tseng, G., Becker-Reshef, I., 2024. How accurate are existing land cover maps for agriculture in Sub-Saharan Africa? Sci. Data 11, 486. https://doi.org/10.1038/s41597-024-03306-z

Khan, H., Ahmad, A., 2025. Evaluating AlphaEarth Foundation Embeddings for Pixel- and Object-Based Land Cover Classification in Google Earth Engine. https://doi.org/10.20944/preprints202511.2172.v1

Klemmer, K., Rolf, E., Robinson, C., Mackey, L., Rußwurm, M., 2025. SatCLIP: Global, General-Purpose Location Embeddings with Satellite Imagery. Proc. AAAI Conf. Artif. Intell. 39, 4347–4355. https://doi.org/10.1609/aaai.v39i4.32457

Klemmer, K., Rolf, E., Rußwurm, M., Camps-Valls, G., Czerkawski, M., Ermon, S., Francis, A., Jacobs, N., Kerner, H., Mackey, L., Mai, G., Mac Aodha, O., Reichstein, M., Robinson, C., Rolnick, D., Shelhamer, E., Sitzmann, V., Tuia, D., Zhu, X., 2026. Earth Embeddings: Toward artificial intelligence-centric representations of our planet. IEEE Geosci. Remote Sens. Mag. 2–15. https://doi.org/10.1109/MGRS.2026.3710416

Ko Ko, H.Y., Atzberger, C., 2026. Exploring the potential of AlphaEarth and TESSERA embeddings for Fine-scale Local Climate Zone Mapping: A case study across five cities in Switzerland. https://doi.org/10.48550/arXiv.2606.20034

Koch, J., Denager, T., Stisen, S., Greve, M.H., Møller, A.B., Beucher, A.M., 2026. Modelling Carbon and Groundwater in Peatlands using AlphaEarth Embeddings. https://doi.org/10.22541/essoar.177135853.32181864/v2

Kumari, S., Agarwal, S., Agrawal, N.K., Agarwal, A., Garg, M.C., 2025. A Comprehensive Review of Remote Sensing Technologies for Improved Geological Disaster Management. Geol. J. 60, 223–235. https://doi.org/10.1002/gj.5072

Lane, K., Karimzadeh, M., 2026. A Genealogy of Foundation Models in Remote Sensing. ACM Trans. Spat. Algorithms Syst. 12, 15:1–15:36. https://doi.org/10.1145/3789505

Li, K., Cao, X., Meng, D., 2024. A New Learning Paradigm for Foundation Model-Based Remote-Sensing Change Detection. IEEE Trans. Geosci. Remote Sens. 62, 5610112. https://doi.org/10.1109/TGRS.2024.3365825

Li, M., Qi, J., Ye, S., Wang, Q., 2025. Manifold embeddings achieve comparable performance with multispectral imagery for time-series based land disturbance detection. Int. J. Digit. Earth 18, 2523481. https://doi.org/10.1080/17538947.2025.2523481

Liang, M., Duncanson, L., Silva, J.A., Sedano, F., 2023. Quantifying aboveground biomass dynamics from charcoal degradation in Mozambique using GEDI Lidar and Landsat. Remote Sens. Environ. 284, 113367. https://doi.org/10.1016/j.rse.2022.113367

Liang, M., Minor, D., Silva, J., Fatoyinbo, L., Thomas, N., van Goor, W., Riemer, K., Kigenyi, R., Cohen, R., Mekuria, A., Williams, M., Marshall, A., Field, C., Jin, C., Healey, S., Cook-Patton, S., Costedoat, S., Heilmayr, R., Chazdon, R., Hansen, M., Dubayah, R., Urbazaev, M., Durrant, E., de Haas, S., Duncanson, L., 2026. Climate Mitigation Benefits of Forest Restoration Increase Over Time in East Africa. https://doi.org/10.21203/rs.3.rs-10361711/v1

Liu, Q., Huang, T., Dong, Y., Yang, J., Xiang, W., 2025. From Pixels to Images: Deep Learning Advances in Remote Sensing Image Semantic Segmentation. https://doi.org/10.48550/arXiv.2505.15147

Liu, W., Li, S., Shi, C., Zhu, H., Huang, C., Yin, L., 2026. Landslide Mapping and Susceptibility Assessment in the Middle and Lower Reaches of the Nujiang River (2017–2025) Using Satellite Embedding and Multidimensional Environmental Factors. Remote Sens. 18, 1854. https://doi.org/10.3390/rs18111854

Lobell, D.B., Di Tommaso, S., Zhou, Q., Ma, Y., Specht, J., Guan, K., 2025. The mixed effects of recent cover crop adoption on US cropland productivity. Nat. Sustain. 8, 1004–1012. https://doi.org/10.1038/s41893-025-01599-5

Lu, Chris, Lu, Cong, Lange, R.T., Foerster, J., Clune, J., Ha, D., 2024. The AI Scientist: Towards Fully Automated Open-Ended Scientific Discovery. https://doi.org/10.48550/arXiv.2408.06292

Ma, Y., Zhang, Z., Kang, Y., Özdoğan, M., 2021. Corn yield prediction and uncertainty analysis based on remotely sensed variables using a Bayesian neural network approach. Remote Sens. Environ. 259, 112408. https://doi.org/10.1016/j.rse.2021.112408

Ma, Y., Liang, S.-Z., Myers, D.B., Swatantran, A., Lobell, D.B., 2024. Subfield-level crop yield mapping without ground truth data: A scale transfer framework. Remote Sens. Environ. 315, 114427. https://doi.org/10.1016/j.rse.2024.114427

Ma, Y., Shen, Y., Swatantran, A., Lobell, D.B., 2026. Harvesting AlphaEarth: Benchmarking the geospatial foundation model for agricultural downstream tasks. Int. J. Appl. Earth Obs. Geoinformation 149, 105258. https://doi.org/10.1016/j.jag.2026.105258

Mahato, T., Sarkar, S., Parida, B.R., Bar, S., 2026. Beyond Multispectral Limits: Satellite Embeddings For Fire Severity Estimation Across Global Biomes. https://doi.org/10.21203/rs.3.rs-9849270/v1

Marsocci, V., Jia, Y., Le Bellier, G., Kerekes, D., Zeng, L., Hafner, S., Gerard, S., Brune, E., Yadav, R., Shibli, A., Fang, H., Ban, Y., Vergauwen, M., Audebert, N., Nascetti, A., 2024. PANGAEA: A Global and Inclusive Benchmark for Geospatial Foundation Models. https://doi.org/10.48550/arXiv.2412.04204

Murakami, K., 2025. Within- and cross-regional crop classification for cool climate upland agriculture using AlphaEarth. https://doi.org/10.31220/agriRxiv.2025.00354

Nakata, N., Liu, J., Chen, G., Nakata, R., Varadharajan, C., 2026. Subsurface Property Mapping using Google AlphaEarth Foundations. https://doi.org/10.48550/arXiv.2604.14756

Oquab, M., Darcet, T., Moutakanni, T., Vo, H., Szafraniec, M., Khalidov, V., Fernandez, P., Haziza, D., Massa, F., El-Nouby, A., Assran, M., Ballas, N., Galuba, W., Howes, R., Huang, P.-Y., Li, S.-W., Misra, I., Rabbat, M., Sharma, V., Synnaeve, G., Xu, H., Jegou, H., Mairal, J., Labatut, P., Joulin, A., Bojanowski, P., 2024. DINOv2: Learning Robust Visual Features without Supervision. Trans. Mach. Learn. Res. https://openreview.net/forum?id=a68SUt6zFt

Ou, Z., Zheng, Y., 2026. Foundation-Scale Satellite Embeddings Reframe Hydrological Generalization as a Representation Problem. Geophys. Res. Lett. 53, e2025GL121604. https://doi.org/10.1029/2025GL121604

Pascual, A., Guerra-Hernández, J., 2026. Integration of Google's Alpha Earth Foundations into biomass estimation combined with GEDI spaceborne lidar and field inventory data. For. Ecol. Manag. 606, 123550. https://doi.org/10.1016/j.foreco.2026.123550

Peng, D., Liu, M., Zhang, Y., Guan, H., 2025. Toward Label-Efficient Deep Learning Change Detection for Remote Sensing Imagery: A Comprehensive Review. Photogramm. Rec. 40, e70021. https://doi.org/10.1111/phor.70021

Pettersson, M.B., Daoud, A., 2025. Leveraging Compact Satellite Embeddings and Graph Neural Networks for Large-Scale Poverty Mapping. https://doi.org/10.48550/arXiv.2511.01408

Ronneberger, O., Fischer, P., Brox, T., 2015. U-Net: Convolutional Networks for Biomedical Image Segmentation, in: Navab, N., Hornegger, J., Wells, W.M., Frangi, A.F. (Eds.), Medical Image Computing and Computer-Assisted Intervention – MICCAI 2015. Springer International Publishing, Cham, pp. 234–241. https://doi.org/10.1007/978-3-319-24574-4_28

Roy, D.P., Wulder, M.A., Loveland, T.R., Woodcock, C.E., Allen, R.G., Anderson, M.C., Helder, D., Irons, J.R., Johnson, D.M., Kennedy, R., Scambos, T.A., Schaaf, C.B., Schott, J.R., Sheng, Y., Vermote, E.F., Belward, A.S., Bindschadler, R., Cohen, W.B., Gao, F., Hipple, J.D., Hostert, P., Huntington, J., Justice, C.O., Kilic, A., Kovalskyy, V., Lee, Z.P., Lymburner, L., Masek, J.G., McCorkel, J., Shuai, Y., Trezza, R., Vogelmann, J., Wynne, R.H., Zhu, Z., 2014. Landsat-8: Science and product vision for terrestrial global change research. Remote Sens. Environ. 145, 154–172. https://doi.org/10.1016/j.rse.2014.02.001

Ryan, S., Powell, M., Ling, J., Wen, L., 2026. Streamlining Wetland Vegetation Mapping with AlphaEarth Embeddings: Comparable Accuracy to Traditional Methods with Cleaner Maps and Minimal Preprocessing. Remote Sens. 18, 293. https://doi.org/10.3390/rs18020293

Salas, J., Zea-Ortiz, M., Vera, P., Wood, D., 2026. A Two-Stage Approach to Improve Poverty Mapping Spatial Resolution. Remote Sens. 18, 427. https://doi.org/10.3390/rs18030427

Seydi, S.T., 2025. Deep Learning-Based Burned Area Mapping Using Bi-Temporal Siamese Networks and AlphaEarth Foundation Datasets. https://doi.org/10.48550/arXiv.2509.07852

Sohrabi, H., 2026. Integrating satellite embeddings and national forest inventory data to map above-ground biomass of temperate deciduous old-growth forests. Adv. Space Res. 77, 11837–11850. https://doi.org/10.1016/j.asr.2026.04.018

Stewart, A.J., Fang, H., Corley, I.A., Zhu, X.X., 2026. Earth Embeddings. https://doi.org/10.48550/arXiv.2608.03410

Timoner, P., Poussin, C., Feng, Z., Planque, C., Chognard, S., Chatenoux, B., Lambiel, A., Lucas, R., Giuliani, G., 2026. Exploring the Potential of Geo-Embeddings to Produce Relevant Environmental Descriptors for Land Cover Mapping. https://papers.ssrn.com/sol3/papers.cfm?abstract_id=6331733

Tong, X.-Y., Wang, S., 2025. Invariant Features for Global Crop Type Classification. https://doi.org/10.48550/arXiv.2509.03497

Tseng, G., Fuller, A., Reil, M., Herzog, H., Beukema, P., Bastani, F., Green, J.R., Shelhamer, E., Kerner, H., Rolnick, D., 2025. Galileo: Learning Global & Local Features of Many Remote Sensing Modalities, in: Proceedings of the 42nd International Conference on Machine Learning. Presented at the International Conference on Machine Learning, PMLR, pp. 60280–60300.

U.S. Department of Agriculture, 2024. National Agriculture Imagery Program (NAIP). https://doi.org/10.5066/F7QN651G

Vivanco Cepeda, V., Nayak, G.K., Shah, M., 2023. GeoCLIP: Clip-Inspired Alignment between Locations and Images for Effective Worldwide Geo-localization. Adv. Neural Inf. Process. Syst. 36, 8690–8701. https://doi.org/10.52202/075280-0379

Wang, Y., Xiong, Z., Liu, C., Stewart, A.J., Dujardin, T., Bountos, N.I., Zavras, A., Gerken, F., Papoutsis, I., Leal-Taixé, L., Zhu, X.X., 2025. Towards a Unified Copernicus Foundation Model for Earth Vision, in: 2025 IEEE/CVF International Conference on Computer Vision (ICCV). Presented at the 2025 IEEE/CVF International Conference on Computer Vision (ICCV), pp. 9888–9899. https://doi.org/10.1109/ICCV51701.2025.00922

Wellmann, T., Lausch, A., Andersson, E., Knapp, S., Cortinovis, C., Jache, J., Scheuer, S., Kremer, P., Mascarenhas, A., Kraemer, R., Haase, A., Schug, F., Haase, D., 2020. Remote sensing in urban planning: Contributions towards ecologically sound policies? Landsc. Urban Plan. 204, 103921. https://doi.org/10.1016/j.landurbplan.2020.103921

Wulder, M.A., Roy, D.P., Radeloff, V.C., Loveland, T.R., Anderson, M.C., Johnson, D.M., Healey, S., Zhu, Z., Scambos, T.A., Pahlevan, N., Hansen, M., Gorelick, N., Crawford, C.J., Masek, J.G., Hermosilla, T., White, J.C., Belward, A.S., Schaaf, C., Woodcock, C.E., Huntington, J.L., Lymburner, L., Hostert, P., Gao, F., Lyapustin, A., Pekel, J.-F., Strobl, P., Cook, B.D., 2022. Fifty years of Landsat science and impacts. Remote Sens. Environ. 280, 113195. https://doi.org/10.1016/j.rse.2022.113195

Xu, Y., Ma, Y., Zhang, Z., 2024. Self-supervised pre-training for large-scale crop mapping using Sentinel-2 time series. ISPRS J. Photogramm. Remote Sens. 207, 312–325. https://doi.org/10.1016/j.isprsjprs.2023.12.005

Yang, J., Du, B., Zhang, L., 2023. From center to surrounding: An interactive learning framework for hyperspectral image classification. ISPRS J. Photogramm. Remote Sens. 197, 145–166. https://doi.org/10.1016/j.isprsjprs.2023.01.024

Yang, J., Du, B., Zhang, L., 2024. Overcoming the Barrier of Incompleteness: A Hyperspectral Image Classification Full Model. IEEE Trans. Neural Netw. Learn. Syst. 35, 14467–14481. https://doi.org/10.1109/TNNLS.2023.3279377

Yang, J., Du, B., Liu, R., Mao, Z., Zhang, L., 2026a. Boosting Multimodal Remote Sensing Image Classification With Transformer-Based Heterogeneously Salient Graph Representation. IEEE Trans. Geosci. Remote Sens. 64, 5512117. https://doi.org/10.1109/TGRS.2026.3686762

Yang, L., Gao, Y., Zhao, X., Liang, N., Ma, R., Xi, S., Zhang, X., Wang, R., 2026b. Evaluating the Performance of AlphaEarth Foundation Embeddings for Irrigated Cropland Mapping Across Regions and Years. Remote Sens. 18, 1065. https://doi.org/10.3390/rs18071065

Yao, T., Zhang, Y., Qiu, Z., Pan, Y., Mei, T., 2021. SeCo: Exploring Sequence Supervision for Unsupervised Representation Learning. Proc. AAAI Conf. Artif. Intell. 35, 10656–10664. https://doi.org/10.1609/aaai.v35i12.17274

Yu, Z., Li, T., Zhu, Y., Pan, R., 2024. Exploring Foundation Models in Remote Sensing Image Change Detection: A Comprehensive Survey. https://doi.org/10.48550/arXiv.2410.07824

Zhang, B., Wang, H., Alabri, A., Bot, K., McCall, C., Hamilton, D., Růžička, V., 2022. Unsupervised Wildfire Change Detection based on Contrastive Learning. https://doi.org/10.48550/arXiv.2211.14654

Zhang, C., Kerner, H., Wang, S., Hao, P., Li, Z., Hunt, K.A., Abernethy, J., Zhao, H., Gao, F., Di, L., Guo, C., Liu, Z., Yang, Z., Mueller, R., Boryan, C., Chen, Q., Beeson, P.C., Zhang, H.K., Shen, Y., 2025. Remote sensing for crop mapping: A perspective on current and future crop-specific land cover data products. Remote Sens. Environ. 330, 114995. https://doi.org/10.1016/j.rse.2025.114995

Zhu, X.X., Xiong, Z., Wang, Y., Stewart, A.J., Heidler, K., Wang, Y., Yuan, Z., Dujardin, T., Xu, Q., Shi, Y., 2026. On the foundations of Earth foundation models. Commun. Earth Environ. 7, 103. https://doi.org/10.1038/s43247-025-03127-x